\pdfoutput=1

\documentclass[11pt,hyphens]{article}
\usepackage{natbib}
\usepackage[hypcap=false,labelfont=bf]{caption} 
\usepackage{acl} 
\usepackage{times} 
\usepackage{latexsym} 
\usepackage[T1]{fontenc} 
\usepackage[utf8]{inputenc} 
\usepackage{microtype} 
\usepackage{amsmath} 
\usepackage{multirow} 
\usepackage{geometry} 
\usepackage{booktabs, multirow} 
\usepackage{float, stfloats} 
\usepackage{pdflscape} 
\usepackage{longtable} 
\usepackage{caption} 
\usepackage{graphicx} 
\usepackage{enumitem} 
\usepackage{xcolor} 
\usepackage{tikz}
\usepackage{tabularx} 
\usepackage{placeins}
\usepackage{hyperref}
\usepackage{fontawesome5}  

\usetikzlibrary{shapes.geometric, arrows.meta, positioning} 
\usepackage[bottom, multiple]{footmisc} 
\usepackage[most]{tcolorbox} 
\usepackage{hyperref}

\definecolor{darkgreen}{rgb}{0.0, 0.5, 0.0}
\definecolor{darkred}{rgb}{0.6, 0.0, 0.0}
\definecolor{lightgrey}{rgb}{0.95, 0.95, 0.95}

\title{\fontsize{13.8pt}{13pt}\selectfont LSC-Eval: A General Framework to Evaluate Methods for Assessing Dimensions of Lexical Semantic Change Using LLM-Generated Synthetic Data}

\newcommand{\psy}{\ensuremath{^\Psi}}  
\newcommand{\cs}{\ensuremath{^\lambda}}  
\newcommand{\qm}{\ensuremath{^\Phi}} 
\newcommand{\ati}{\ensuremath{^\mathcal{T}}} 
\newcommand{\ltl}{\ensuremath{^\Sigma}} 

\author{
  Naomi Baes\psy, Raphaël Merx\cs, Nick Haslam\psy, Ekaterina Vylomova\cs, Haim Dubossarsky\qm\ati\ltl\\
  \psy Melbourne School of Psychological Sciences, The University of Melbourne \\
  \cs School of Computing and Information Systems, The University of Melbourne \\
  \qm School of Electronic Engineering and Computer Science, Queen Mary University of London \\
  \ati The Alan Turing Institute, London \\
  \ltl Language Technology Lab, University of Cambridge \\
  \footnotesize \texttt{\{n.baes, r.merx, nhaslam, vylomovae\}@unimelb.edu.au, h.dubossarsky@qmul.ac.uk}}

\begin{document}
\maketitle
\begin{abstract} 

Lexical Semantic Change (LSC) provides insight into cultural and social dynamics. Yet, the validity of methods for measuring different kinds of LSC remains unestablished due to the absence of historical benchmark datasets. To address this gap, we propose LSC-Eval, a novel three-stage general-purpose evaluation framework to: (1) develop a scalable methodology for generating synthetic datasets that simulate theory-driven LSC using In-Context Learning and a lexical database; (2) use these datasets to evaluate the sensitivity of computational methods to synthetic change; and (3) assess their suitability for detecting change in specific dimensions and domains. We apply LSC-Eval to simulate changes along the Sentiment, Intensity, and Breadth (SIB) dimensions, as defined in the SIBling framework, using examples from psychology. We then evaluate the ability of selected methods to detect these controlled interventions. Our findings validate the use of synthetic benchmarks, demonstrate that tailored methods effectively detect changes along SIB dimensions, and reveal that a state-of-the-art LSC model faces challenges in detecting affective dimensions of LSC. LSC-Eval offers a valuable tool for dimension- and domain-specific benchmarking of LSC methods, with particular relevance to the social sciences.


\href{https://github.com/naomibaes/LSCD_method_evaluation}{%
\faGithub\ \texttt{https://github.com/\allowbreak naomibaes/\allowbreak LSCD\_method\_evaluation}}




\end{abstract}

\section{Introduction} 


Lexical Semantic Change (LSC) provides a unique window into cultural dynamics by revealing how language evolution reflects social changes \citep{mcgillivray2020computational, schlechtweg-etal-2020-semeval}. Recently developed state-of-the-art (SOTA) computational methods have expanded our ability to classify established types of LSC, such as generalization and specialization \cite{cassotti-etal-2024-using}. Efforts have also been directed towards developing methods for measuring newly proposed dimensions of LSC \cite{Baes:2024,desá2024surveycharLSC}. Nevertheless, the field faces challenges in validating these methods. A major obstacle is the absence of historical benchmark datasets, which restricts the standardization and fair comparison of metrics. Additionally, there is a pressing need for fine-grained evaluation methods that save  time and resources.

To address these challenges, the present study proposes LSC-Eval, a novel three-stage evaluation framework for assessing LSC across dimensions and domains. It: (1) develops a scalable, general-purpose methodology for generating high-quality synthetic sentences using `scholar-in-the-loop' In-Context Learning (ICL) and a lexical resource to simulate targeted kinds of LSC; (2) uses these datasets to evaluate the validity of alternative LSC detection methods; and (3) identifies the most suitable methods for capturing change in specific dimensions and domains. We apply LSC-Eval to test the sensitivity of various methods to synthetically induced changes along three major LSC dimensions — Sentiment, Intensity, and Breadth (SIB) — as proposed in the SIBling framework (\citealt{Baes:2024}; see Table~\ref{tab:semantic_dimensions}), drawing from psychology examples. Illustrative cases include \textit{awesome}, which has risen in sentiment (from serious to enthusiastic), declined in intensity (from awe-inspiring to casually pleasing), and broadened in scope (from solemn reverence to general positivity); \textit{trauma}, which has shifted from physical injury to a broader and milder range of psychological harms; and \textit{sick}, which has risen in sentiment (in slang), broadened in usage (from a medical term to a general expression of praise), and declined in intensity (from a descriptor of serious illness to one used in casual or enthusiastic contexts).

\vspace{-5pt}  

\begin{table*}[ht]
\setlength{\abovecaptionskip}{5pt} 
\setlength{\belowcaptionskip}{5pt} 
\centering
\scriptsize 
\renewcommand{\arraystretch}{1.2} 
\begin{tabular}{@{}p{1cm} p{3.6cm} p{4.9cm} p{4.9cm}@{}}
\toprule
\textbf{Dimension} & \textbf{Definition} & \textbf{Examples of Rising} & \textbf{Examples of Falling} \\ \midrule

\textit{\textbf{Sentiment}} & 
Relates to the degree to which a word's meaning acquires more positive (\textit{`elevation', `amelioration'}) or negative (\textit{`degeneration', `pejoration'}) connotations. & 
\textit{craftsman}, once associated with manual labor, has come to convey artistry, skill, and high-quality workmanship.
\vspace{2pt}
\newline \textit{geek}, originally a derogatory term for odd people, now refers to someone passionate about a field. & 
\textit{retarded}, originally a neutral term for intellectual disability, has become highly pejorative over time. 
\vspace{2pt}
\newline 
\textit{awful}, once meaning "awe-inspiring," now indicates something very bad. \\ \midrule

\textit{\textbf{Intensity}} & 
Relates to the degree to which a word’s meaning changes to acquire more (\textit{`meiosis'}) or less (\textit{`hyperbole'}) emotionally charged (i.e., strong, potent, high-arousal) connotations. & 
\textit{cool} has evolved from describing temperature to expressing strong approval or trendiness.
\vspace{2pt}
\newline \textit{hilarious}, originally meaning cheerful or amusing in Latin, has come to describe extremely funny things that cause great merriment and laughter. & 

\textit{love} has evolved from denoting romantic or platonic attachment to also expressing milder forms of liking (e.g., “I love coffee.”).
\vspace{2pt}
\newline 
\textit{trauma} has shifted from referring to brain injuries to encompassing milder events (e.g., business loss). \\ \midrule

\textit{\textbf{Breadth}} & 
Relates to the degree to which a word expands (\textit{`widening', `generalization'}) or contracts (\textit{`narrowing', `specialization'}) its semantic range. & 
\textit{cloud}, initially a meteorological term, broadened in usage to refer to internet-based data storage.
\vspace{2pt}
\newline 
\textit{partner}, which once referred narrowly to a business associate, has broadened to denote a romantic or domestic companion. & 
\textit{doctor}, once used for any scholar or teacher, now primarily denotes a medical professional.
\vspace{2pt}
\newline 
\textit{meat}, which originally referred to any kind of food in Old English (\textit{mete}), has since narrowed to denote animal flesh as food. \\

\bottomrule
\end{tabular}
\caption{Definitions and Examples of \citeauthor{Baes:2024}'s (\citeyear{Baes:2024}) Dimensions of Lexical Semantic Change.}
\label{tab:semantic_dimensions}
\end{table*}

The present study addresses two key research questions: (RQ1) Can synthetic datasets be used to validate methods for measuring LSC dimensions? We hypothesize that SIB scores will be associated with levels of induced change; (RQ2) Which of several LSC detection methods is most sensitive to synthetically induced changes in SIB? Our findings confirm the validity of theory-driven changes — that is, changes systematically injected into corpora based on the SIBling framework \cite{Baes:2024}, which targets the dimensions of Sentiment, Intensity, and Breadth. These changes were operationalized using ICL and a lexical resource to simulate diachronic semantic shifts. Results highlight the need to tailor detection methods to specific dimensions, as the state-of-the-art LSC model failed to detect affective change. The proposed framework, LSC-Eval, offers an efficient and scalable approach for dimension- and domain-specific benchmarking of LSC methods. While broadly applicable, LSC-Eval is especially valuable in the social sciences and humanities, where capturing nuanced conceptual change is essential.

\section{Related Work}

\subsection{Theoretical Background}

Linguists have long debated taxonomies of LSC \cite{breal1900semantics,Blank:1999}, defined as innovations which change word meanings \cite{bloomfield1933}. A growing body of work has identified ways to detect changes in the meanings of words and quantify the extent of these changes using computational approaches \citep{kutuzov-etal-2018-diachronic, tahmasebi2018survey, tang2018state, tahmasebi2023computational, cassotti-etal-2024-computational, Periti_Montanelli2024, kiyama-etal-2025-analyzing}. 

Recent years have seen the development of theoretical frameworks that propose multiple dimensions of LSC. \citet{Baes:2024} introduced SIBling, a three-dimensional framework that maps LSC along axes of SIB, reflecting a word's acquisition of more positive or negative connotations (Sentiment), more or less emotionally charged or potent connotations (Intensity), and the expansion or contraction of its semantic range (Breadth). It draws on linguistic \cite{Geeraerts:2010} and psychological \citep{Haslam2016} theories, and provides methodological tools to estimate SIB across time. In parallel, \citet{desá2024surveycharLSC} proposed a framework that clusters LSC into three dimensions using graph structures: Orientation (shifts towards more pejorative or ameliorated senses), Relation (changes towards metaphoric or metonymic usage), and Dimension (variations between abstract/general and specific/narrow meanings). While \citet{desá2024surveycharLSC} surveyed statistical methods for representing word meaning (word frequency, topic modeling, and graph structures) on dimensions, they did not demonstrate their usage.

Both frameworks contain dimensions of evaluation (Sentiment and Orientation) and semantic range (Breadth and Dimension). Baes et al.'s \citeyearpar{Baes:2024} inclusion of Intensity reflects a greater emphasis on changes in the emotional connotations of words. Sentiment and Intensity resemble the two primary dimensions of human emotion, Valence and Arousal \cite{Russell:2003}, and two primary dimensions of connotational meaning, Evaluation (e.g., “good/bad”) and Potency (e.g., “strong/weak”) \cite{Osgood:1975}, which have been demonstrated to have cross-cultural validity. 


\subsection{Evaluation}

Despite substantial progress in developing benchmarks \cite{tahmasebi-risse-2017-finding} and evaluation strategies \cite{kutuzov-etal-2018-diachronic}, the field still lacks standardized datasets that evaluate multiple dimensions of LSC across time. Current annotated benchmarks, such as the synchronic, definition- and type-based \textit{LSC Cause-Type-Definitions Benchmark} \citep{cassotti-etal-2024-using} and the binary, word-sense-based \textit{TempoWIC}, where LSC is labeled by comparing the sameness or difference of meanings between two sense usages \citep{loureiro-etal-2022-tempowic}, address different aspects of semantic change.


The first human-annotated dataset of LSC in multiple languages (English, German, Latin, Swedish; \citealp{schlechtweg-etal-2020-semeval}) marked substantial progress in identifying the presence and degree of LSC but omitted information about kinds of change. However, expert-annotated datasets are costly and time-intensive to create. To address this gap, \citet{Dubossarsky:2019} introduced a method to artificially induce semantic change in controlled experimental settings, enabling precise testing of how well models capture these shifts.

Recent developments in generative artificial intelligence highlight the potential of pre-trained LLMs to adapt to novel tasks at inference time through ICL \cite{zhou2023mystery}. Few-shot ICL, a paradigm that enables LLMs to learn tasks by analogy given only a few demonstrative examples, helps to incorporate theoretical knowledge without needing to fine-tune its internal parameters \cite{dong-etal-2024-survey}. Instead, ICL uses context from the model's prompt to adapt the LLM to downstream tasks \cite{Radford2019LanguageMA, brown2020language, liu2024}. \citet{de2024semantic} demonstrated the utility of few-shot ICL, employing Chain-of-Thought and rhetorical devices, to annotate LSC dimensions, but their strategy focused on multi-class classification of change between two sense usages. ICL offers a promising solution to bridge the absence of standardized approaches \citep{hengchen_2021} for assessing the effectiveness of different methods to measure dimensions of LSC.



\section{Method}

\subsection{Materials}

\subsubsection{Psychology Corpus}
\label{subsec:psych_corpus}  

To develop and test the evaluation pipeline on a specific domain, a corpus of psychology article abstracts was sourced \citep{Vylomova:2019}. It includes 133,017,962 tokens from 871,337 abstracts (1970-2019) from E-Research and PubMed databases, and contains 5,214,227 sentences.\footnote{Sentences were segmented using "en\_core\_web\_sm" (\url{https://spacy.io/models/en}); F-score = 91\%.} 

\subsubsection{WordNet}
\label{subsec:wordnet}  

Although other ontologies were considered,\footnote{PsycNET, UMLS, DSM-5, ConceptNet} the English WordNet lexical database 3.0 \cite{miller-1992-wordnet} was chosen for its linguistic coverage and lexical structure. It organizes words into synsets, linking them by semantic relationships (e.g., hypernyms, hyponyms). 

\subsubsection{Targets}
\label{subsec:targets}  

While the evaluation framework is general in its applicability, six terms from psychology — \textit{abuse}, \textit{anxiety}, \textit{depression}, \textit{mental health}, \textit{mental illness}, and \textit{trauma} — were analyzed for semantic change, selected for their empirical and theoretical relevance to shifting word meanings. \textit{Trauma}, \textit{mental health}, and \textit{mental illness} have seen falls in their average valence alongside their semantic expansions (\textit{trauma}: \citealp{Baes2023, haslam2021cultural}; \textit{mental health}, \textit{mental illness}: \citealp{Baes:2024}). There have been changes in their semantic intensity, with rises for \textit{mental health} and \textit{mental illness} \citep{Baes:2024}, as well as \textit{anxiety} and \textit{depression} \citep{xiao2023}, and a fall for \textit{trauma} \citep{Baes2023}. Qualitatively, \textit{abuse} has expanded horizontally to include passive neglect and emotional abuse, beyond its physical scope \cite{Haslam2016}. Targets were sufficiently prevalent (sentence counts: 46,272; 104,486; 115,430; 44,130; 5,808; 23,187). Appendix~\ref{sec:appendix_A} shows annual counts. 

\subsection{Evaluation Framework}
\label{sec:eval}

The three general stages of the evaluation framework are illustrated in Figure \ref{fig:pipeline}. 

\tikzstyle{arrow} = [thick,-Stealth]

\begin{figure}[H]
    \centering
    \begin{tikzpicture}[node distance=1.5cm and 0cm]
    \node (step1) [rectangle, rounded corners, minimum width=2.5cm, minimum height=0.1cm, text centered, draw=black!75, fill=gray!20] {\textbf{Stage 1:} Generate and Validate Synthetic Datasets};
    \node (step2) [rectangle, rounded corners, minimum width=2.5cm, minimum height=0.1cm, text centered, draw=black!75, fill=orange!30, below=0.5cm of step1] {\textbf{Stage 2:} Evaluate the Effectiveness of Methods};  
    \node (step3) [rectangle, rounded corners, minimum width=2.5cm, minimum height=0.1cm, text centered, draw=black!75, fill=green!25, below=0.5cm of step2] {\textbf{Stage 3:} Select the Best-Performing Method}; 
    \draw [arrow] (step1) -- (step2);
    \draw [arrow] (step2) -- (step3);
    \end{tikzpicture}
    \caption{Stages of the Evaluation Framework.}
    \label{fig:pipeline}
\end{figure}
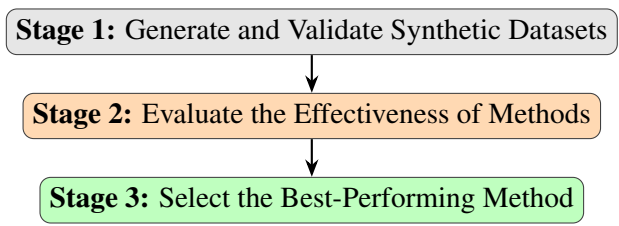

In Stage 1, synthetic datasets (i.e., datasets exhibiting simulated variations in linear semantic change along a specific dimension) were created to benchmark changes in LSC dimensions using few-shot ICL and a lexical database. GPT-4o \cite{achiam2023gpt}\footnote{ChatGPT API documentation: \url{https://platform.openai.com/docs/guides/text-generation}} was prompted with expert-crafted examples to increase and decrease corpus sentences in affective dimensions across five-year intervals. This ensures that synthetic sentences are theory-driven, domain-specific and contain temporal features. GPT was used due to its adeptness at few-shot learning, task adaptation with minimal examples \citep{achiam2023gpt, merx-etal-generating-examples} and lack of disciplinary bias \cite{ziems-etal-2024-large}. Table~\ref{tab:synthetic_data_descriptives} provides descriptive statistics for the synthetic datasets,\footnote{Link to Synthetic datasets: \url{https://github.com/naomibaes/Synthetic-LSC_pipeline}} which are validated using tools that have been shown to measure SIB in historical (domain-general and -specific) corpora \cite{Baes:2024}. 

\begin{table}[!ht]
\centering
\resizebox{\columnwidth}{!}{%
\begin{tabular}{@{}llrrrr@{}}
\toprule
\textbf{Dimension} & \textbf{Target} & \textbf{Neutral (\textit{M})} & \textbf{Increase (\textit{M})} & \textbf{Decrease (\textit{M})} & \textbf{US\$} \\
\midrule
\multirow{6}{*}{\textbf{\textit{Sentiment}}} 
                                & \textit{abuse} & 5,645 (28) & 5,645 (30) & 5,645 (29) & 17 \\
                                & \textit{anxiety} & 9,215 (27) & 9,213 (28) & 9,213 (28) & 28 \\
                                & \textit{depression} & 8,828 (27) & 8,826 (28) & 8,826 (28) & 29 \\
                                & \textit{mental health} & 6,348 (28) & 6,348 (29) & 6,348 (29) & 21 \\
                                & \textit{mental illness} & 2,552 (28) & 2,552 (28) & 2,552 (29) & 9 \\
                                & \textit{trauma} & 3,563 (28) & 3,563 (30) & 3,563 (30) & 11 \\
\midrule
\multirow{6}{*}{\textbf{\textit{Intensity}}} 
                                & \textit{abuse} & 6,802 (28) & 6,801 (30) & 6,801 (29) & 21 \\
                                & \textit{anxiety} & 9,659 (26) & 9,657 (29) & 9,657 (28) & 32 \\
                                & \textit{depression} & 10,022 (27) & 10,020 (30) & 10,020 (29) & 35 \\
                                & \textit{mental health} & 6,904 (28) & 6,899 (32) & 6,899 (29) & 24 \\
                                & \textit{mental illness} & 2,497 (28) & 2,496 (32) & 2,496 (29) & 10 \\
                                & \textit{trauma} & 4,012 (28) & 4,012 (30) & 4,012 (30) & 14 \\
\midrule
\multirow{6}{*}{\textbf{\textit{Breadth}}} 
                                & \textit{abuse} & -- & 5,221 (27) & -- & 0 \\
                                & \textit{anxiety} & -- & 13,635 (26) & -- & 0 \\
                                & \textit{depression} & -- & 14,463 (27) & -- & 0 \\
                                & \textit{mental health} & -- & 14,638 (26) & -- & 0 \\
                                & \textit{mental illness} & -- & 14,639 (26) & -- & 0 \\
                                & \textit{trauma} & -- & 14,650 (26) & -- & 0 \\
\bottomrule
\end{tabular}
}
\caption{Descriptive Statistics for Synthetic Dimension Datasets: Sentence Counts, Mean Lengths, and Total Generation Cost.}
\caption*{\textit{Note.} \textit{M} = Mean sentence word length. Neutral = Original input sentences. Increase = Sentences modified to increase the dimension of interest. Decrease = Sentences modified to decrease the dimension of interest.}

\label{tab:synthetic_data_descriptives}
\end{table}

\paragraph{Experimental Settings:}

We implemented two sampling strategies to assess the sensitivity of LSC detection methods. \textit{Bootstrap sampling} involved randomly selecting 50 sentences with replacement from the full corpus (natural or synthetic), repeated over 100 iterations. This allowed any sentence to be sampled multiple times, enabling robust estimation through resampling. \textit{Five-year interval sampling} selected up to 50 unique sentences per time bin, repeated 10 times. Sentences were not repeated within an iteration but could appear across iterations. This approach ensured balanced temporal coverage and better reflects natural language variability over time. Each sampling strategy had a \textit{control condition}, where sentences were shuffled to balance natural and synthetic ones per sample, providing a baseline to isolate the effect of synthetic injection, following prior work in computational linguistics \citep{dubossarsky-etal-2017-outta, Dubossarsky:2019}.

\begin{figure}[h!] 
\centering 
\includegraphics[width=\linewidth]{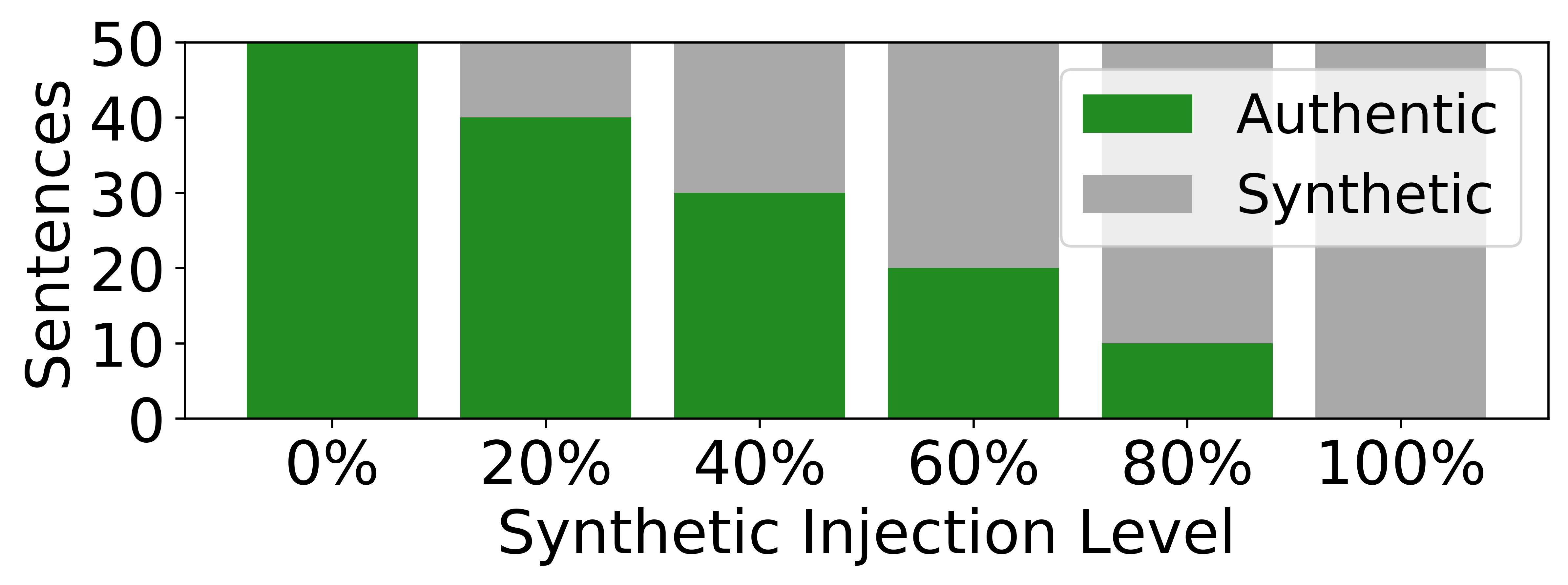} 
\caption{Distribution of Sentences in Each Sample.} 
\label{fig:injection_levels} 
\end{figure}

For each sampling strategy, synthetic sentences were injected into (up to) 50-sentence samples at increasing proportions: 20\%, 40\%, 60\%, 80\%, and 100\% (see Figure~\ref{fig:injection_levels}). These injection levels simulate increasing semantic saturation to test method sensitivity (Stage~2). For method selection (Stage~3), only the 0\% and 100\% injection levels were used. This enables the clearest contrast between unchanged (fully natural) and fully altered (synthetic) samples, maximizing the signal-to-noise ratio when evaluating which method detects the greatest magnitude of semantic change.

\vspace{-5pt} 

\subsection{Synthetic Dataset Generation}

For each semantic dimension and target (Section~\ref{subsec:targets}), we generated datasets by inducing change in natural sentences drawn from the corpus (Section~\ref{subsec:psych_corpus}). Each dataset included up to 1,500 randomly sampled sentences per five-year interval used in the experimental settings (see Section~\ref{sec:eval}). See Appendix~\ref{sec:appendix_B} for examples of synthetic sentences for each dimension and target.  

\subsubsection{Sentiment and Intensity}

To generate the synthetic Sentiment and Intensity datasets, we employed few-shot ICL with GPT-4o to vary these dimensions. First, neutral diachronic sentences from the corpus (see Section~\ref{subsec:psych_corpus}) were sampled as outlined in Appendix~\ref{sec:appendix_C}. Second, a psychology scholar crafted five \cite{chen-etal-2023-many} diverse sentence variations for each target, to serve as few-shot demonstrations for the LLM in the prompt outlined below. This prompt includes construct definitions, a task description, and few-shot examples to generate theory-driven change — that is, changes aligned with the dimensions in the SIBling framework (e.g., shifts from positive to negative sentiment or from high to low intensity). For `scholar-in-the-loop' few-shot demonstrations, see Appendix~\ref{sec:appendix_D} (Sentiment), Appendix~\ref{sec:appendix_E} (Intensity), and the few-shot demonstration examples below. 

\begin{tcolorbox}[sharp corners, colframe=black!75, colback=white!95, boxrule=0.5mm, title=\textbf{Prompt Outline for Synthetic \textit{Sentiment}}]
\small
\textbf{Prompt intro:} In psychology research, `Sentiment' is defined as "a term's acquisition of a more positive or negative connotation." This task focuses on the sentiment of the term \textbf{\texttt{target\_word}}. \\
\textbf{Task:} You will be given a sentence containing the term \textbf{\texttt{target\_word}}. Your goal is to write two new sentences:
\begin{enumerate}[itemsep=1pt, parsep=1pt, topsep=2pt, partopsep=0pt]
    \item One where \textbf{\texttt{target\_word}} has a \textbf{more positive connotation}.
    \item One where \textbf{\texttt{target\_word}} has a \textbf{more negative connotation}.
\end{enumerate}    
\textbf{Guidelines:}  \textit{[Rules and important notes to constrain model output and make it contextually realistic.]} \\
--------------------------------------------------------------- \\
\vspace{2pt} 
\textbf{Append few-shot examples:} \textit{[One example below.]}
\begin{tcolorbox}[colback=gray!20, colframe=gray!20, sharp corners, boxrule=0.5mm, left=2pt, right=2pt, top=2pt, bottom=2pt]
\textcolor{darkgray}{\textbf{Neutral Sentence}:} Previous work suggests that social \textit{\textbf{anxiety}} is inconsistently related to alcohol use. \\
\textcolor{darkgreen}{\textbf{Positive Variation}:} Previous work \underline{agrees} that social \textit{\textbf{anxiety}} is \underline{sometimes} related to alcohol use. \\
\textcolor{darkred}{\textbf{Negative Variation}:} Previous work \underline{warns} that social \textit{\textbf{anxiety}} is \underline{unpredictably} related to alcohol use. 
\end{tcolorbox}
\end{tcolorbox}

\begin{tcolorbox}[sharp corners, colframe=black!75, colback=white!95, boxrule=0.5mm, title=\textbf{Prompt Outline for Synthetic \textit{Intensity}}]
\small
\textbf{Prompt intro:} In psychology research, `Intensity' is defined as "the degree to which a word has emotionally charged (i.e., strong, potent, high-arousal) connotations." This task focuses on the intensity of the term \textbf{\texttt{target\_word}}. \\
\textbf{Task:} You will be given a sentence containing the term \textbf{\texttt{target\_word}}. Your goal is to write two new sentences:
\begin{enumerate}[itemsep=1pt, parsep=1pt, topsep=2pt, partopsep=0pt]
    \item One where \textbf{\texttt{target\_word}} is \textbf{less intense}.
    \item One where \textbf{\texttt{target\_word}} is \textbf{more intense}.
\end{enumerate}    
\textbf{Guidelines:}  \textit{[Rules and important notes to constrain model output and make it contextually realistic.]} \\
--------------------------------------------------------------- \\
\vspace{2pt} 
\textbf{Append few-shot examples:} \textit{[One example below.]}
\begin{tcolorbox}[colback=gray!20, colframe=gray!20, sharp corners, boxrule=0.5mm, left=2pt, right=2pt, top=2pt, bottom=2pt]
\textcolor{darkgray}{\textbf{Neutral sentence}:} They tend to be more liberal in their attitudes toward abortion than women in general; however, women who experienced a greater degree of psychic \textit{\textbf{trauma}} tended to be more conservative in their attitudes. \\
\textcolor{darkred}{\textbf{Low Variation}:} They tend to be more \underline{accepting} in their attitudes toward \underline{children} than women in general; however, women who experienced \underline{mild} psychic \textit{\textbf{trauma}} tended to be more conservative in their attitudes.

\textcolor{darkgreen}{\textbf{High Variation}:} They tend to be more \underline{extremely callous} in their attitudes toward \underline{the horrors of} abortion than women in general; however, women who \underline{suffered} a greater degree of \underline{violent} psychic \textit{\textbf{trauma}} tended to be more \underline{fearful} in their attitudes. \\
\end{tcolorbox}
\end{tcolorbox}

Third, the prompt was refined during pilot tests (varying 10 input sentences for each target). Fourth, for each of the neutral sentences (Sentiment: 36,151; Intensity: 39,896), we made one inference call to GPT-4o through the OpenAI API to generate variations of Sentiment (positive/negative) or Intensity (high/low). Fifth, validation checks required checking if output sentences retained the target term. A few manual adjustments were made by a tertiary-educated English native speaker to ensure the final sentences retained the target in a similar location to the original input sentence. Few output sentences required manual adjustment in the final datasets: 0.25\% for the synthetic Sentiment dataset, and 0.01\% for the synthetic Intensity dataset. See Table~\ref{tab:synthetic_data_descriptives} for the counts of input/output sentences and final prompts (dataset costs: for Sentiment = 115 \$US; for Intensity = 136 \$US). 

\subsubsection{Breadth}

Unlike Sentiment and Intensity, current Breadth measures have no score that assigns a mid-point with which to obtain neutral sentences to vary. Therefore, to induce semantic breadth, we adapted \citeauthor{Dubossarsky:2019}'s (\citeyear{Dubossarsky:2019}) replacement strategy, using WordNet 3.0 (see Section \ref{subsec:wordnet}) to expand a target word’s usage by incorporating contexts from donor terms, thereby broadening its semantic range without altering its core meaning. Relevant synsets were identified and filtered for psychological relevance using keyword matching\footnote{Psychology key terms: "abnormality", "abnormally", "emotional", "feeling", "feelings", "harm", "hurt","mental", "mind", "psychological", "psychology", "psychiatry", "syndrome", "therapy", "treatment".} and semantic similarity thresholds. Donor terms (co-hyponyms with the target) were filtered using Lin similarity (0.5)\footnote{Information content values from the psychology corpus.} and cosine similarity (0.7) using embeddings from BioBERT \cite{lee2020biobert}, a pre-trained language model for biomedical text mining, to capture context-dependent meanings of synset glosses in 768-dimensional vectors. See Appendix~\ref{sec:appendix_F} for the full list of siblings (i.e., co-hyponyms). The sibling replacement process identifies and replaces sibling terms with the target, as illustrated below. To sample representatively from the sibling list, a round-robin strategy was used, where the algorithm runs through each sibling in the list in batch sizes, randomly sampling up to 50 unique sentences before proceeding to the next one. In this way, it exhaustively samples up to 1,500 unique sentences per epoch and injection level to create the final synthetic breadth dataset.

\begin{tcolorbox}[sharp corners, colframe=black!75, colback=white!95, boxrule=0.5mm, title=\textbf{Dataset Creation for Synthetic \textit{Breadth} }]
\small
\textbf{Replacement Strategy:} Randomly sample sentences containing co-hyponyms of the target term from the validated list and replace the \textbf{\texttt{co-hyponym}} with the \textbf{\texttt{target}} to be used as a synthetic sentence. \\  
--------------------------------------------------------------- \\
\vspace{2pt} 
\textit{[One example for \textbf{mental\_health} below.]}
\begin{tcolorbox}[colback=gray!20, colframe=gray!20, sharp corners, boxrule=0.5mm, left=2pt, right=2pt, top=2pt, bottom=2pt]
\textcolor{darkgray}{\textbf{Donor Context}:} The `Angry and Impulsive Child' and 'Abandoned and Abused Child' modes uniquely predicted  \textit{\textbf{dissociation}} scores. \\
\textcolor{darkgreen}{\textbf{Synthetic Context}:} The `Angry and Impulsive Child' and 'Abandoned and Abused Child' modes uniquely predicted  \underline{\textit{\textbf{mental health}}} scores.
\end{tcolorbox}
\end{tcolorbox}

\subsection{Quantifying Lexical Semantic Change}
\subsubsection{Semantic Dimensions}

We applied methodologies from \citet{Baes:2024} to assess changes in semantic Sentiment, Intensity, and Breadth. Sentiment and Intensity were quantified by assigning ordinal valence and arousal scores, based on \citet{Warriner:2013} norms, matched to collocates within ±5 words of the target, using a scale (ranging from 1-9) from `extremely unhappy' to `extremely happy' for valence, and `extremely calm' to `extremely agitated' for arousal. Scores were frequency-weighted, normalized, and averaged across bins, yielding indices from 0 (negative/low arousal) to 1 (positive/high arousal). Breadth was measured by averaging the cosine distances between sentence-level embeddings from the SentenceTransformer model `all-mpnet-base-v2'\footnote{Microsoft pretrained network (109M model params) \url{https://huggingface.co/sentence-transformers/all-mpnet-base-v2}} (MPNet). The Breadth score indicates semantic range variation from 0 (no change) to 1 (maximum variation). See Appendix~\ref{sec:appendix_G1} for further details.

For comparative analysis, Sentiment was compared against the Deberta-v3-base-absa-v1.1\footnote{yangheng/deberta-v3-base-absa-v1.1 (184M model params): \url{https://huggingface.co/yangheng/deberta-v3-base-absa-v1.1}} aspect-based sentiment analysis (ABSA) classification model, which we adapted to output continuous sentiment scores ranging from 0 (fully negative) to 1 (fully positive). Because Intensity is a novel dimension, it lacks comparable methods. For the Breadth comparisons, we used \citeauthor{Cassotti:2023}'s (\citeyear{Cassotti:2023}) SOTA \cite{periti-tahmasebi-2024-systematic} word transformer "XL-LEXEME"\footnote{XL-LEXEME (\textasciitilde550M model params) \url{https://huggingface.co/pierluigic/xl-lexeme}} (XLL). While MPNet generates sentence embeddings through pooling tokens, which dilutes word-specific information, XLL uses a bi-encoder architecture that focuses on word-specific attention, using polysemy as a proxy for meaning divergence during training. See Appendix~\ref{sec:appendix_G2} for further details on the quantification of semantic Breadth.


\subsubsection{General Lexical Semantic Change}

To quantify general LSC, we used the SOTA LSC score \citep{Cassotti:2023}, which calculates the Average Pairwise Cosine Distance (APD) between sentence embeddings from two time periods, following the approach of \citet{giulianelli-etal-2020-analysing}. We extended this method to compare embeddings from different bins within the same iteration. See Appendix~\ref{sec:appendix_G3} for more details.


\subsection{Statistical Strategy}

To test whether synthetic change influences SIB scores (RQ1), we fit separate mixed linear models for each outcome variable (Valence, Arousal, and Breadth), with random intercepts for target terms and fixed effects for injection level (see Appendix~\ref{sec:appendix_H} for model specifications). To assess the sensitivity of various computational methods to detecting LSC (RQ2), we calculated the percent relative change in a target word’s context between the 0\% (fully natural) and 100\% (fully synthetic) injection levels, as defined in Equation~\ref{eq:percent_change}:

{\small
\begin{equation}
\Delta\% = \frac{X_{100} - X_{0}}{X_{0}} \times 100
\label{eq:percent_change}
\end{equation}
}

Here, \(X_0\) and \(X_{100}\) denote the score at 0\% and 100\% injection, respectively. This index quantifies how much the contextual meaning of the target word shifts when all natural sentences are replaced with synthetic ones.

For the XLL LSC Score, we further normalized this change to account for internal variability within each bin, as shown in Equation~\ref{eq:normalized_change}. We used the APD, the mean cosine distance between all sentence pairs,\footnote{APD reflects semantic dispersion; higher APD implies greater variability.} to control for within-bin noise:

{\small
\begin{equation}
\Delta = \frac{\text{APD}(X_{100}\text{-between-}X_{0})}
{\max[\text{APD}(X_{0}\text{-within-}X_{0}), \text{APD}(X_{100}\text{-within-}X_{100})]}
\label{eq:normalized_change}
\end{equation}
}

The numerator reflects divergence between bins, while the denominator accounts for within-bin variability, ensuring that the score reflects genuine LSC signal rather than internal fluctuation.

\section{Results}

\begin{figure}[!ht] 
\setlength{\abovecaptionskip}{4pt} 
\setlength{\belowcaptionskip}{4pt} 
    \centering
    \captionsetup{position=below}
    \includegraphics[width=0.48\textwidth]{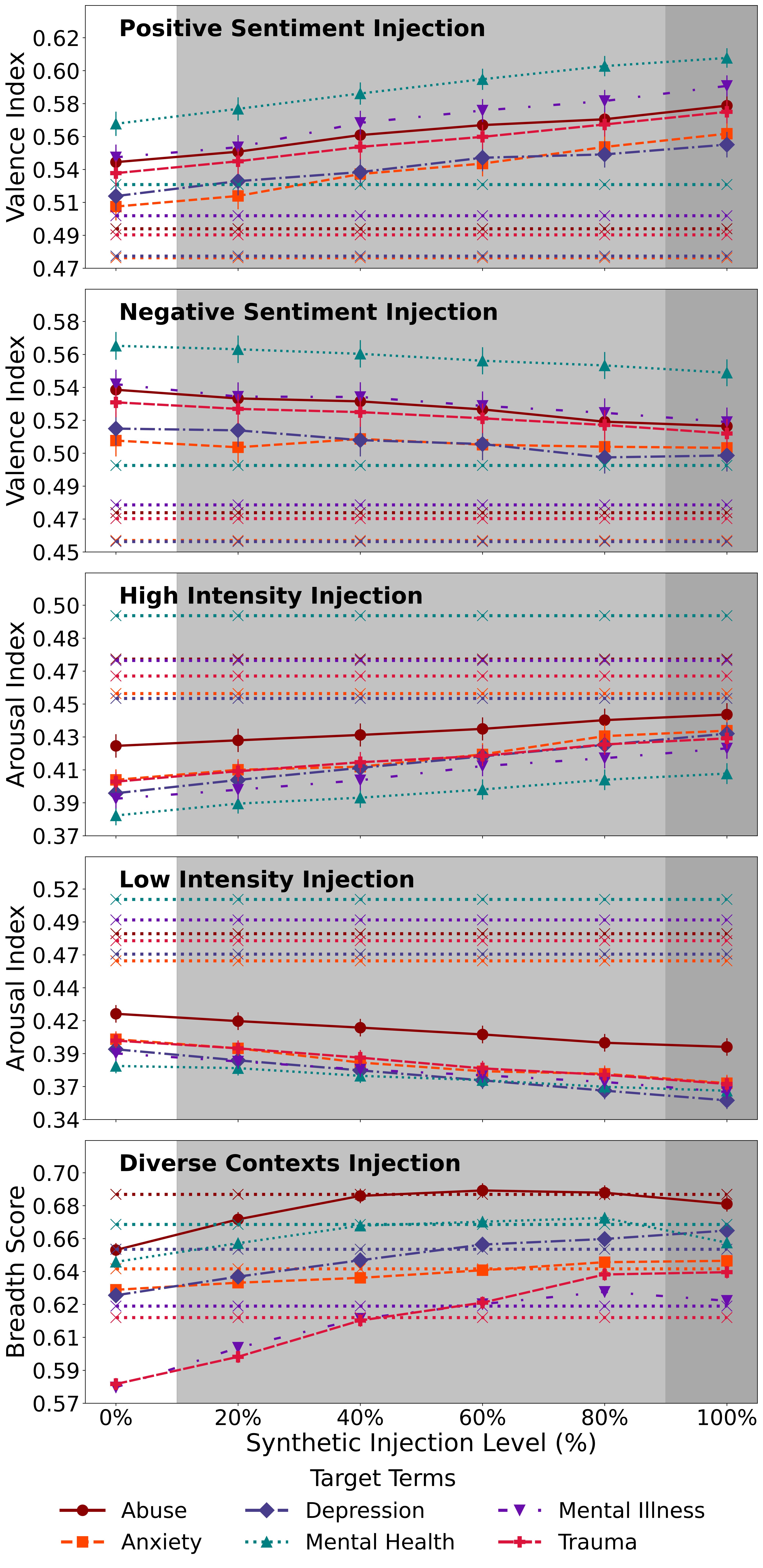}
    \caption{SIB Scores (±SE) by Injection Levels for Experimental and Control Settings (Flat Dotted Lines).}
    \label{fig:3}
\end{figure}
\begin{figure}[ht] 
\setlength{\abovecaptionskip}{4pt} 
\setlength{\belowcaptionskip}{4pt} 
    \centering
    \captionsetup{position=below}
    \includegraphics[width=0.48\textwidth]{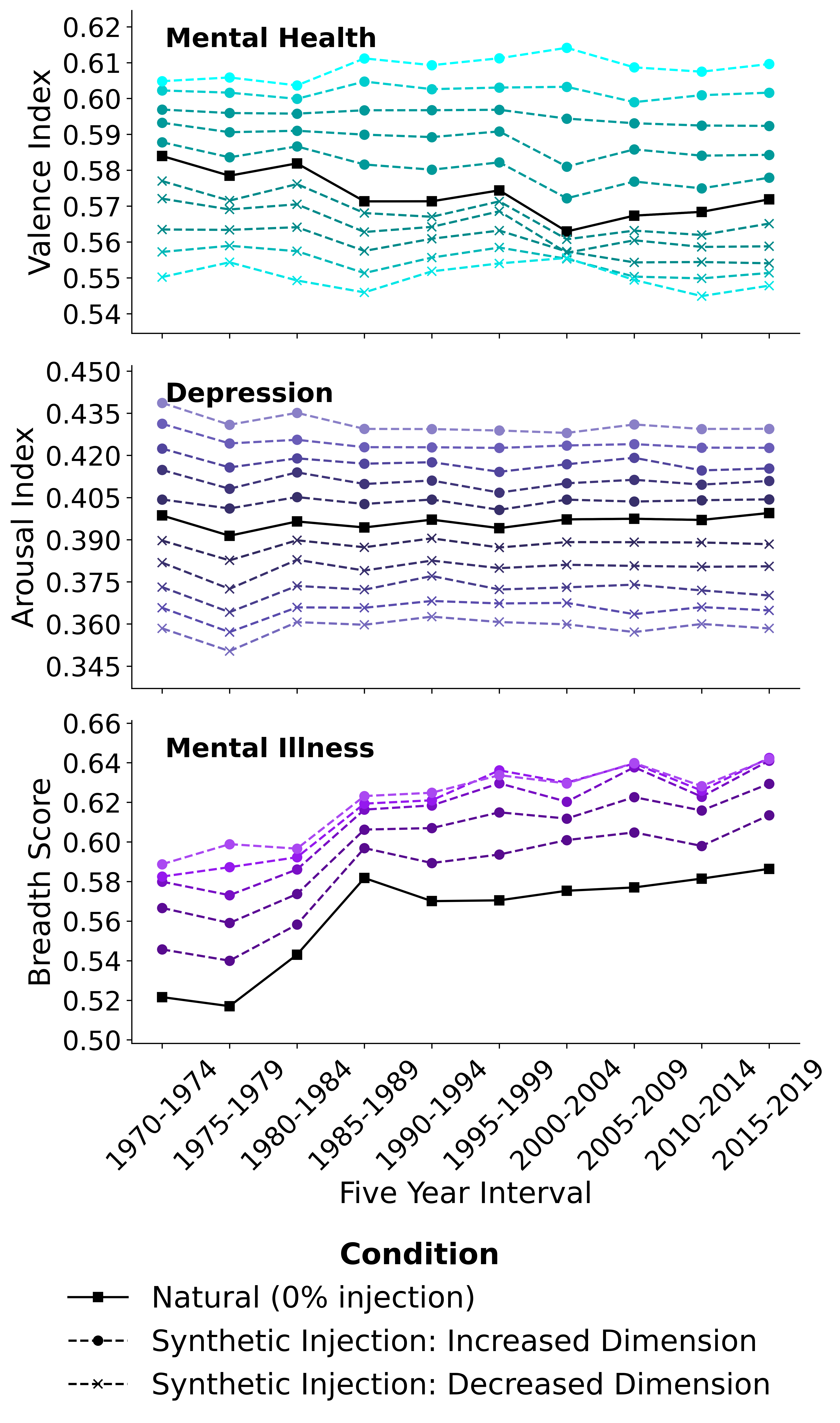}
    \caption{SIB scores by five-year intervals across injection levels and conditions.}
    \label{fig:4} 
\end{figure}

\paragraph{Synthetic Change Effects:} 
To address RQ1, we tested whether SIB scores from \citeauthor{Baes:2024}'s (\citeyear{Baes:2024}) SIBling framework are associated with levels of synthetically induced change. The hypothesis was supported: standardized SIB scores across all three dimensions — Sentiment, Intensity, and Breadth — demonstrated consistent increases or decreases as a function of injection level, across targets and both sampling strategies: bootstrap (Figure~\ref{fig:3}) and five-year intervals (Figure~\ref{fig:4}).


Figure~\ref{fig:4} illustrates SIB scores by five-year intervals across injection levels and conditions for three of the six targets. The black line in each panel depicts the changes in SIB scores over time in the natural corpus, while the colored lines represent injection levels. The intensity of the color reflects the percentage of synthetic sentences injected into each sample (20\% to 100\%). As intended, injection levels altered the height, but not necessarily the slope, of SIB scores on the \textit{y}-axis, indicating changes in the magnitude of the score rather than trend over time. For breadth (bottom panel), only upward (broadening) changes were modeled. 

Mixed-effects linear models confirmed that a one-standard-deviation increase in injection level significantly predicted changes in SIB scores in the intended direction (Table~\ref{tab:MLM_maintext}; see also Appendix~\ref{sec:appendix_H}).

\begin{table}[H]
\centering
\begin{tabular}{@{}lccc@{}}
\toprule
\textbf{Score} & \textbf{Valence} & \textbf{Arousal} & \textbf{Breadth} \\
\midrule
\( \beta^+ \) & .61* & .64* & .43* \\
\( \beta^- \) & $-$.31* & $-$.64* & -- \\
\bottomrule
\end{tabular}
\caption[Short caption for LoT]{Coefficients of random intercept (by target) models predicting SIB scores from injection level.}
\label{tab:MLM_maintext}
\end{table}



\paragraph{Control Experiments:} As illustrated in Figure~\ref{fig:3}, controlling for synthetic injection level by re-analyzing data with shuffled sentences for uniform distribution revealed flat SIB score trends in bootstrapped settings. Appendix~\ref{sec:appendix_I} illustrated how, in the control condition for the five-year interval setting, SIB scores in those samples tend to converge to a midpoint between natural and synthetic data.

\paragraph{Comparative Method Evaluation:} 
To address RQ2 — identifying which methods are most sensitive to synthetically induced changes in SIB — we compared the performance of alternative change detection methods across each dimension. Results were mixed. For Sentiment, both the Valence index and ABSA’s Sentiment score were sensitive to detecting synthetic change, though ABSA outperformed the Valence index in 10/12 cases. For Intensity, the Arousal index shows sensitivity to detecting variations in synthetic Intensity. For Breadth, XLL outperformed MPNet in 4/6 cases using the Breadth score.
 
\begin{figure}[ht] 
\setlength{\abovecaptionskip}{4pt} 
\setlength{\belowcaptionskip}{4pt} 
    \centering
    \captionsetup{position=below}
    \includegraphics[width=0.48\textwidth]{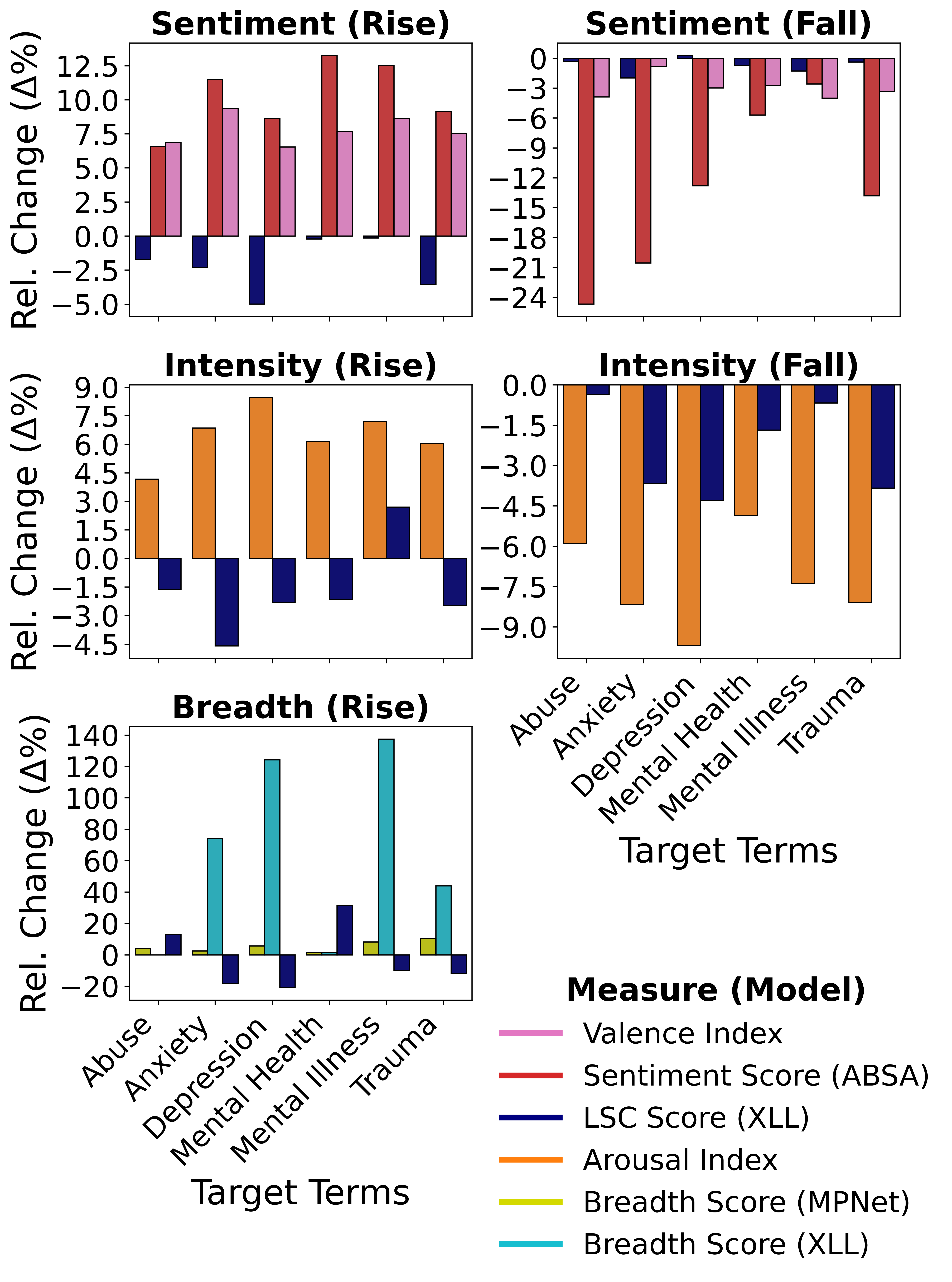}
    \caption{Relative Change (\(\Delta\%\)) Scores for Models Across Dimensions and Conditions: Bootstrapped.}
    \label{fig:plot_method_comparison_all-year}
\end{figure}

Critically, XLL-LSC score is completely insensitive to detecting changes in either Sentiment or Intensity. XLL-LSC can only indicate change via positive change values, while negative values indicate that the within-bin variance is greater than the change scores between bins. See Appendix~\ref{sec:appendix_J} for between- and within-bin LSC scores across \textit{all} synthetic injection levels. Thus, the negative scores observed in Sentiment and Intensity (except for \textit{mental illness}) establish that XLL was unable to detect any change signal in these words. XLL-LSC detects changes in synthetic Breadth for 2/6 terms.


\section{Discussion}

The present study introduced LSC-Eval, a three-stage general evaluation framework that: (1) creates synthetic datasets featuring scholar-in-the-loop LLM-generated sentences to simulate various kinds of LSC; (2) leverages these datasets to assess the sensitivity of computational approaches to synthetic change; and (3) evaluates the suitability of these methods for specific dimensions and domains. We applied this framework to generate synthetic datasets that induce changes across the three dimensions (Sentiment, Intensity, and Breadth) of a novel multidimensional LSC framework, SIBling \cite{Baes:2024}, using examples drawn from psychology to compare the suitability of alternative methods for detecting these synthetic changes.

Importantly, these synthetic datasets reflect theory-driven change, operationalized through construct definitions and distinctions drawn from the SIBling framework \citep{Baes:2024}, which integrates established traditions in lexical semantics, connotation, and affect. This principled approach enables interpretable and targeted benchmarking, and lays the groundwork for generalizing LSC-Eval to other theoretical models and domains.

Our findings support the hypothesis that recently proposed methods (Valence index, Arousal index, Breadth score; \citealp{Baes:2024}) detect synthetic changes on the SIB (i.e., Sentiment, Intensity, and Breadth) dimensions. This directly addresses RQ1, which asked whether synthetic datasets can be used to validate methods for measuring LSC dimensions. Control analyses, which adhered to computational linguistics standards \cite{dubossarsky-etal-2017-outta, Dubossarsky:2019}, confirmed the absence of these effects in shuffled controls. The implications of these findings are two-fold. First, the ability of SIB methods to detect changes introduced via silver-standard synthetic data validates their sensitivity and reliability for detecting and quantifying variation in SIB, even in artificial environments. Second, it confirms the validity of using LLM-generated sentences in our ICL evaluation suites.

To address RQ2 — identifying which computational methods are most sensitive to synthetically induced changes in SIB — we demonstrated how a synthetic change detection task can assess the sensitivity of various approaches, guiding the selection of the most suitable method for specific dimensions and domains. \citeauthor{Baes:2024}'s (\citeyear{Baes:2024}) tools, which validated the synthetic SIB datasets, were supported by alternative methods that consistently detected synthetic changes across conditions and targets, providing further validation. The Valence index and ABSA's Sentiment score detected variations in synthetic Sentiment, while Breadth scores derived from XLL and MPNet models detected increases in synthetic Breadth. These results suggest that the NLP-based ABSA method is generally more sensitive to synthetically induced change than Warriner-based approaches, which rely on static Valence and Arousal ratings. Future empirical studies on Sentiment and Breadth may consider adopting these NLP-based models, either as replacements for, or in addition to, traditional lexicon-based methods.


Notably, when computing the general LSC score using the SOTA LSC model XLL \cite{Cassotti:2023}, it was not sensitive to detecting Sentiment and Intensity. Although XLL shows some sensitivity to identifying synthetic increases in Breadth, it registers a more substantial change when the Breadth score is adjusted according to the method introduced by \citet{Baes:2024}. It uses the within-bin average cosine distance of target-containing sentences as a proxy for the expansion (broadening) or contraction (narrowing) of a word's contextual usage. The inability of XLL to detect the affective dimensions of LSC highlights the necessity of evaluating SOTA models before deploying them in new domains. Future research should investigate whether this weakness in detecting affective dimensions is specific to XLL or extends to other contextualized models in other corpora and concepts. This inquiry is particularly salient given recent advances in analyzing fine-grained, continuous semantic shifts through ``diachronic word similarity matrices using fast and lightweight word embeddings over arbitrary time periods" \cite{kiyama-etal-2025-analyzing}. Our results support the idea that current SOTA methods are only state-of-the-art with respect to a particular kind of change or a specific domain and should not be assumed to generalize beyond them.

Findings highlight the need to include affective and connotational aspects of meaning in studies of LSC. In particular, future studies must consider emotional meaning in language models. While psychology has extensively used language to analyze emotion semantics \cite{Jackson2022, boyd2021natural}, advances in NLP are still exploring how to build models that incorporate sentiment \cite{goworek-dubossarsky-2024-toward} and detect emotion \cite{mohammad2021sentiment}. Further research is required to detect affective states from text given the cultural and universal aspects of emotion semantics \cite{jackson2019}. These findings also have implications for existing multidimensional frameworks of LSC \cite{Baes:2024, desá2024surveycharLSC} as the evaluation framework provides experimental settings in which to compare the sensitivity of methods to detecting synthetic changes on specific dimensions and domains in a variety of disciplines.

\section{Conclusion}

The current study introduced a novel general-purpose evaluation framework, LSC-Eval, for evaluating methods that detect LSC. Its three-stage approach: 1) generates LLM-based synthetic datasets with silver-standard labels that simulate changes in kinds of LSC; 2) uses these datasets to evaluate the relative sensitivity of computational methods in a synthetic change detection task; and 3) identifies the most suitable method for detecting synthetically induced changes across specific dimensions and domains. We applied this framework to a set of psychological terms. Findings not only supported the validity of proposed computational methods for measuring changes in SIB, but also established a controlled experimental standard for rigorously evaluating existing LSC detection methods and exploring alternative computational approaches. This work is crucial for addressing the substantial gap created by the lack of historical benchmark datasets, which has previously hindered the standardization of metrics and fair comparison of methods. While this framework benefits all disciplines (e.g., biomedicine, law, theology), LSC-Eval is particularly valuable in the social sciences and humanities, where domain-specific constructs require specialized tools for capturing nuanced semantic change.

\newpage

\section*{Limitations}

Limitations inform future directions. Evaluating the quality of LLM prompt and demonstration examples in the few-shot ICL paradigm is challenging. As LLM evaluation standards are developed \cite{Wang2024, ziems-etal-2024-large}, future research might explore automated strategies such as updating prompts based on examples (DSPy)\footnote{\url{https://dspy.ai}} or comparing LLM output from different prompts using a free, unified interface.\footnote{\url{https://github.com/marketplace/models/azure-openai/gpt-4o/playground}} LLM choice in the evaluation pipeline could be expanded to include open-source models (e.g., FlanT5-XL, Mistral-7B, Mixtral-8x7B). Future work could also compare this approach to a non-generative rule-based one. 

Furthermore, our study benefited from using GPT-4o, which is trained on US English and is therefore well-suited for analyzing texts within the Western-centric domain of psychology \cite{varnum2024large}. However, cultural and linguistic biases of LLMs can pose challenges in adapting our evaluation pipeline to other languages \cite{havaldar-etal-2023-multilingual}, although few-shot ICL has proven effective in low-resource languages \cite{cahyawijaya-etal-2024-llms}. Despite the tendency of LLM training data to skew towards the recent past, manual inspection of results demonstrate the successful generation of quality sentences that spanned a 1970 to 2019 time period. Future work should focus on refining these models to broadly apply across cultural contexts, languages, and historical periods.

The conceptualization of semantic Breadth is complex and contested. Linguistic definitions suggest breadth encompasses subtypes (e.g., specialization as a subtype of narrowing; \citealp{Campbell2013}) highlighting its intricate nature. Given this complexity, it is essential to compare the current measure, which is based on mean within-bin variability of target-containing sentences, with other methods assessing breadth through senses, topics, or prototypical changes: modulations based on literal similarity \cite{geeraerts1997diachronic}. Future research should investigate whether these measures can detect polysemy's emergence or merely prototype-based modulations of existing concepts.

The synthetic breadth dataset used in this study was constructed using a replacement strategy that may include contextually irrelevant donor contexts. To enhance simulation quality, we propose a three-step validation pipeline: First, select validation models based on performance against a gold-standard dataset, as determined by the highest F1-score from 5-fold stratified K-fold cross-validation. Second, use a probability ratio check with a Masked Language Model (e.g., BioBERT, RoBERTa-large, DeBERTa-v3-large) to confirm the plausibility of replacing donors with target terms, approving sentences that meet a specific probability threshold. Third, ensure semantic alignment through cosine similarity validation with models such as MiniLM-L12-v2 or DistilRoBERTa-v1 Sentence-T5, approving sentences that exceed a set threshold. This process aims to expand the target term’s semantic scope while maintaining specificity, but may exclude many sentences. Integrating \citeauthor{desá2024surveycharLSC}'s (\citeyear{desá2024surveycharLSC}) ICL approach to simulate Breadth—first teaching the model to disambiguate word senses—could offer an efficient alternative.


Furthermore, the present study does not specify which sense of the term is semantically expanded. Attempting to integrate senses into the synthetic data generation pipeline may provide richer insights. While the specialized psychology corpus and target words exhibit limited senses, general domain corpora introduce ambiguous contexts (e.g., the economic sense of ``depression"). Further research is needed on sense-specific LSC. 

Although a body of work estimates valence from natural language, less research has examined the Intensity dimension \cite{hoemannconstruction}. In the present study, this restricted the external validation of the Arousal index \cite{Baes:2024}, highlighting the need for empirical research in this direction. Furthermore, we must examine the conceptual/terminological link between arousal and hyperbole (i.e., a linguistic form describing a rhetorical, discursive phenomenon like irony) to understand arousal's relation to hyperbole \cite{burgers2016figurative, pena2017construing}. 

Finally, future research should use the evaluation framework to generate synthetic datasets, and to explore methods, for detecting the Relation dimension (metaphor/metonymy) as highlighted by \citet{de2024semantic}. Incorporating the qualitative types of metaphor and metonymy into the empirical study of multidimensional LSC could provide a more comprehensive understanding of LSC, particularly for some domains. Examining how Relation relates to SIB may deepen our understanding of LSC processes by exploring how cognitive principles contribute to semantic innovations.


\section*{Ethical Considerations}

We do not foresee any risks or potential for harmful use arising from our research. Our analyses utilize sentences from a psychology corpus, which consists of licensed data openly accessible for academic use, thereby ensuring both transparency and accountability.

\section*{Acknowledgments}


We express our gratitude to the individuals who provided valuable feedback during the early stages of this work: Assistant Professor Ehsan Shareghi for his insightful comments; Professor Emeritus Dirk Geeraerts for discussions on the transparency of LLMs and a multidimensional approach to semantic change, including the qualitative dimensions of metaphor and metonymy; Professor Mark Steedman for discussion surrounding the semantic capabilities of LLMs; and Dr Dominik Schlechtweg for his contributions to our understanding of metaphor and metonymy through cognitive theories of similarity and contiguity. 

Special thanks go to Philip Baes for his consistent support and insightful discussions on methodological challenges. We also appreciate the discussions with Roksana Goworek about the LSC score and XL-LEXEME, and Pierluigi Cassotti, Francesco Periti, and Jader Martins Camboim de Sá for enriching our project's context through their work on semantic change.

We acknowledge the support of the University of Melbourne’s general-purpose High Performance Computing system, Spartan \cite{lafayette2016spartan}, which provided the computational resources necessary for efficient embedding encoding using transformer models. We also thank Jeremy Silver from the Statistical Consulting Centre for his valuable feedback on the random effects modelling.

This research was supported by an Australian Government Research Training Program Scholarship and funded, in part, by Australian Research Council Discovery Project DP210103984 and the research program ``Change is Key!", supported by Riksbankens Jubileumsfond (under reference number M21-0021).

\bibliography{custom}

\begin{thebibliography}{65}
\providecommand{\natexlab}[1]{#1}

\bibitem[{Achiam et~al.(2023)Achiam, Adler, Agarwal, Ahmad, Akkaya, Aleman, Almeida, Altenschmidt, Altman, Anadkat et~al.}]{achiam2023gpt}
Josh Achiam, Steven Adler, Sandhini Agarwal, Lama Ahmad, Ilge Akkaya, Florencia~Leoni Aleman, Diogo Almeida, Janko Altenschmidt, Sam Altman, Shyamal Anadkat, et~al. 2023.
\newblock \href {https://doi.org/10.48550/arXiv.2303.08774} {Gpt-4 technical report}.
\newblock \emph{arXiv preprint arXiv:2303.08774}.

\bibitem[{Baes et~al.(2024)Baes, Haslam, and Vylomova}]{Baes:2024}
Naomi Baes, Nick Haslam, and Ekaterina Vylomova. 2024.
\newblock \href {https://doi.org/10.18653/v1/2024.acl-long.76} {A multidimensional framework for evaluating lexical semantic change with social science applications}.
\newblock In \emph{Proceedings of the 62nd Annual Meeting of the Association for Computational Linguistics (Volume 1: Long Papers)}, pages 1390--1415, Bangkok, Thailand. Association for Computational Linguistics.

\bibitem[{Baes et~al.(2023)Baes, Vylomova, Zyphur, and Haslam}]{Baes2023}
Naomi Baes, Ekaterina Vylomova, Michael Zyphur, and Nick Haslam. 2023.
\newblock \href {https://doi.org/doi:10.58734/plc-2023-0002} {The semantic inflation of “trauma” in psychology}.
\newblock \emph{Psychology of Language and Communication}, 27(1):23--45.

\bibitem[{Blank(1999)}]{Blank:1999}
Andreas Blank. 1999.
\newblock Why do new meanings occur? a cognitive typology of the motivations for lexical semantic change.
\newblock In Andreas Blank and Peter Koch, editors, \emph{Historical semantics and cognition}, pages 61--90. Mouton de Gruter.

\bibitem[{Bloomfield(1933)}]{bloomfield1933}
Leonard Bloomfield. 1933.
\newblock \emph{Language}.
\newblock Compton Printing Works Ltd.

\bibitem[{Boyd and Schwartz(2021)}]{boyd2021natural}
Ryan~L Boyd and H~Andrew Schwartz. 2021.
\newblock \href {https://doi.org/0.1177/0261927X20967028} {Natural language analysis and the psychology of verbal behavior: The past, present, and future states of the field}.
\newblock \emph{Journal of Language and Social Psychology}, 40(1):21--41.

\bibitem[{Br{\'e}al(1900)}]{breal1900semantics}
Michel Br{\'e}al. 1900.
\newblock \emph{Semantics: Studies in the Science of Meaning}.
\newblock WILLIAM HEINEMANN, London.

\bibitem[{Brown et~al.(2020)Brown, Mann, Ryder, Subbiah, Kaplan, Dhariwal, Neelakantan, Shyam, Sastry, Askell, Agarwal, Herbert-Voss, Krueger, Henighan, Child, Ramesh, Ziegler, Wu, Winter, Hesse, Chen, Sigler, Litwin, Gray, Chess, Clark, Berner, McCandlish, Radford, Sutskever, and Amodei}]{brown2020language}
Tom Brown, Benjamin Mann, Nick Ryder, Melanie Subbiah, Jared~D Kaplan, Prafulla Dhariwal, Arvind Neelakantan, Pranav Shyam, Girish Sastry, Amanda Askell, Sandhini Agarwal, Ariel Herbert-Voss, Gretchen Krueger, Tom Henighan, Rewon Child, Aditya Ramesh, Daniel Ziegler, Jeffrey Wu, Clemens Winter, Chris Hesse, Mark Chen, Eric Sigler, Mateusz Litwin, Scott Gray, Benjamin Chess, Jack Clark, Christopher Berner, Sam McCandlish, Alec Radford, Ilya Sutskever, and Dario Amodei. 2020.
\newblock \href {https://proceedings.neurips.cc/paper_files/paper/2020/file/1457c0d6bfcb4967418bfb8ac142f64a-Paper.pdf} {Language models are few-shot learners}.
\newblock In \emph{Advances in Neural Information Processing Systems}, volume~33, pages 1877--1901. Curran Associates, Inc.

\bibitem[{Burgers et~al.(2016)Burgers, Konijn, and Steen}]{burgers2016figurative}
Christian Burgers, Elly~A Konijn, and Gerard~J Steen. 2016.
\newblock \href {https://doi.org/10.1111/comt.12096} {Figurative framing: Shaping public discourse through metaphor, hyperbole, and irony}.
\newblock \emph{Communication theory}, 26(4):410--430.

\bibitem[{Cabello and Akujuobi(2024)}]{cabello-akujuobi-2024-simple}
Laura Cabello and Uchenna Akujuobi. 2024.
\newblock \href {https://doi.org/10.18653/v1/2024.findings-acl.394} {It is simple sometimes: A study on improving aspect-based sentiment analysis performance}.
\newblock In \emph{Findings of the Association for Computational Linguistics: ACL 2024}, pages 6597--6610, Bangkok, Thailand. Association for Computational Linguistics.

\bibitem[{Cahyawijaya et~al.(2024)Cahyawijaya, Lovenia, and Fung}]{cahyawijaya-etal-2024-llms}
Samuel Cahyawijaya, Holy Lovenia, and Pascale Fung. 2024.
\newblock \href {https://doi.org/10.18653/v1/2024.naacl-long.24} {{LLM}s are few-shot in-context low-resource language learners}.
\newblock In \emph{Proceedings of the 2024 Conference of the North American Chapter of the Association for Computational Linguistics: Human Language Technologies (Volume 1: Long Papers)}, pages 405--433, Mexico City, Mexico. Association for Computational Linguistics.

\bibitem[{Campbell(2013)}]{Campbell2013}
Lyle Campbell. 2013.
\newblock \href {http://www.jstor.org/stable/10.3366/j.ctt1g0b5gq} {\emph{Historical Linguistics: An Introduction}}, ned - new edition, 3 edition.
\newblock Edinburgh University Press.

\bibitem[{Cassotti et~al.(2024{\natexlab{a}})Cassotti, De~Pascale, and Tahmasebi}]{cassotti-etal-2024-using}
Pierluigi Cassotti, Stefano De~Pascale, and Nina Tahmasebi. 2024{\natexlab{a}}.
\newblock \href {https://doi.org/10.18653/v1/2024.acl-long.249} {Using synchronic definitions and semantic relations to classify semantic change types}.
\newblock In \emph{Proceedings of the 62nd Annual Meeting of the Association for Computational Linguistics (Volume 1: Long Papers)}, pages 4539--4553, Bangkok, Thailand. Association for Computational Linguistics.

\bibitem[{Cassotti et~al.(2024{\natexlab{b}})Cassotti, Periti, De~Pascale, Dubossarsky, and Tahmasebi}]{cassotti-etal-2024-computational}
Pierluigi Cassotti, Francesco Periti, Stefano De~Pascale, Haim Dubossarsky, and Nina Tahmasebi. 2024{\natexlab{b}}.
\newblock \href {https://aclanthology.org/2024.eacl-tutorials.1/} {Computational modeling of semantic change}.
\newblock In \emph{Proceedings of the 18th Conference of the European Chapter of the Association for Computational Linguistics: Tutorial Abstracts}, pages 1--8, St. Julian{'}s, Malta. Association for Computational Linguistics.

\bibitem[{Cassotti et~al.(2023)Cassotti, Siciliani, DeGemmis, Semeraro, and Basile}]{Cassotti:2023}
Pierluigi Cassotti, Lucia Siciliani, Marco DeGemmis, Giovanni Semeraro, and Pierpaolo Basile. 2023.
\newblock \href {https://doi.org/10.18653/v1/2023.acl-short.135} {Xl-lexeme: Wic pretrained model for cross-lingual lexical semantic change}.
\newblock In \emph{Proceedings of the 61st Annual Meeting of the Association for Computational Linguistics (Volume 2: Short Papers)}, pages 1577--1585, Toronto, Canada. Association for Computational Linguistics.

\bibitem[{Chang et~al.(2024)Chang, Wang, Wang, Wu, Yang, Zhu, Chen, Yi, Wang, Wang, Ye, Zhang, Chang, Yu, Yang, and Xie}]{Wang2024}
Yupeng Chang, Xu~Wang, Jindong Wang, Yuan Wu, Linyi Yang, Kaijie Zhu, Hao Chen, Xiaoyuan Yi, Cunxiang Wang, Yidong Wang, Wei Ye, Yue Zhang, Yi~Chang, Philip~S. Yu, Qiang Yang, and Xing Xie. 2024.
\newblock \href {https://doi.org/10.1145/3641289} {A survey on evaluation of large language models}.
\newblock \emph{ACM Trans. Intell. Syst. Technol.}, 15(3).

\bibitem[{Chen et~al.(2023)Chen, Chen, Zhu, and Zhou}]{chen-etal-2023-many}
Jiuhai Chen, Lichang Chen, Chen Zhu, and Tianyi Zhou. 2023.
\newblock \href {https://doi.org/10.18653/v1/2023.findings-emnlp.745} {How many demonstrations do you need for in-context learning?}
\newblock In \emph{Findings of the Association for Computational Linguistics: EMNLP 2023}, pages 11149--11159, Singapore. Association for Computational Linguistics.

\bibitem[{de~S{\'a} et~al.(2024)de~S{\'a}, Da~Silveira, and Pruski}]{de2024semantic}
Jader Martins~Camboim de~S{\'a}, Marcos Da~Silveira, and C{\'e}dric Pruski. 2024.
\newblock \href {https://arxiv.org/pdf/2407.16624} {Semantic change characterization with llms using rhetorics}.
\newblock \emph{arXiv preprint arXiv:2407.16624}.

\bibitem[{de~Sá et~al.(2024)de~Sá, Silveira, and Pruski}]{desá2024surveycharLSC}
Jader Martins~Camboim de~Sá, Marcos~Da Silveira, and Cédric Pruski. 2024.
\newblock \href {https://arxiv.org/abs/2402.19088} {Survey in characterization of semantic change}.
\newblock \emph{Preprint}, arXiv:2402.19088.

\bibitem[{Dong et~al.(2024)Dong, Li, Dai, Zheng, Ma, Li, Xia, Xu, Wu, Chang, Sun, Li, and Sui}]{dong-etal-2024-survey}
Qingxiu Dong, Lei Li, Damai Dai, Ce~Zheng, Jingyuan Ma, Rui Li, Heming Xia, Jingjing Xu, Zhiyong Wu, Baobao Chang, Xu~Sun, Lei Li, and Zhifang Sui. 2024.
\newblock \href {https://doi.org/10.18653/v1/2024.emnlp-main.64} {A survey on in-context learning}.
\newblock In \emph{Proceedings of the 2024 Conference on Empirical Methods in Natural Language Processing}, pages 1107--1128, Miami, Florida, USA. Association for Computational Linguistics.

\bibitem[{Dubossarsky et~al.(2019)Dubossarsky, Hengchen, Tahmasebi, and Schlechtweg}]{Dubossarsky:2019}
Haim Dubossarsky, Simon Hengchen, Nina Tahmasebi, and Dominik Schlechtweg. 2019.
\newblock \href {https://doi.org/10.18653/v1/P19-1044} {Time-out: Temporal referencing for robust modeling of lexical semantic change}.
\newblock In \emph{Proceedings of the 57th Annual Meeting of the Association for Computational Linguistics}, pages 457--470, Florence, Italy. Association for Computational Linguistics.

\bibitem[{Dubossarsky et~al.(2017)Dubossarsky, Weinshall, and Grossman}]{dubossarsky-etal-2017-outta}
Haim Dubossarsky, Daphna Weinshall, and Eitan Grossman. 2017.
\newblock \href {https://doi.org/10.18653/v1/D17-1118} {Outta control: Laws of semantic change and inherent biases in word representation models}.
\newblock In \emph{Proceedings of the 2017 Conference on Empirical Methods in Natural Language Processing}, pages 1136--1145, Copenhagen, Denmark. Association for Computational Linguistics.

\bibitem[{Geeraerts(1997)}]{geeraerts1997diachronic}
Dirk Geeraerts. 1997.
\newblock \emph{Diachronic Prototype Semantics: A Contribution to Historical Lexicology}.
\newblock Oxford: Clarendon Press.

\bibitem[{Geeraerts(2010)}]{Geeraerts:2010}
Dirk Geeraerts. 2010.
\newblock \emph{Theories of lexical semantics}.
\newblock Oxford University Press.

\bibitem[{Giulianelli et~al.(2020)Giulianelli, Del~Tredici, and Fern{\'a}ndez}]{giulianelli-etal-2020-analysing}
Mario Giulianelli, Marco Del~Tredici, and Raquel Fern{\'a}ndez. 2020.
\newblock \href {https://doi.org/10.18653/v1/2020.acl-main.365} {Analysing lexical semantic change with contextualised word representations}.
\newblock In \emph{Proceedings of the 58th Annual Meeting of the Association for Computational Linguistics}, pages 3960--3973, Online. Association for Computational Linguistics.

\bibitem[{Goworek and Dubossarsky(2024)}]{goworek-dubossarsky-2024-toward}
Roksana Goworek and Haim Dubossarsky. 2024.
\newblock \href {https://aclanthology.org/2024.eacl-srw.28/} {Toward sentiment aware semantic change analysis}.
\newblock In \emph{Proceedings of the 18th Conference of the European Chapter of the Association for Computational Linguistics: Student Research Workshop}, pages 350--357, St. Julian{'}s, Malta. Association for Computational Linguistics.

\bibitem[{Haslam(2016)}]{Haslam2016}
Nick Haslam. 2016.
\newblock \href {https://doi.org/10.1080/1047840X.2016.1082418} {Concept creep: Psychology's expanding concepts of harm and pathology}.
\newblock \emph{Psychological Inquiry}, 27(1):1--17.

\bibitem[{Haslam et~al.(2021)Haslam, Vylomova, Zyphur, and Kashima}]{haslam2021cultural}
Nick Haslam, Ekaterina Vylomova, Michael Zyphur, and Yoshihisa Kashima. 2021.
\newblock \href {https://doi.org/10.1037/amp0000847} {The cultural dynamics of concept creep.}
\newblock \emph{American Psychologist}, 76(6):1013.

\bibitem[{Havaldar et~al.(2023)Havaldar, Singhal, Rai, Liu, Guntuku, and Ungar}]{havaldar-etal-2023-multilingual}
Shreya Havaldar, Bhumika Singhal, Sunny Rai, Langchen Liu, Sharath~Chandra Guntuku, and Lyle Ungar. 2023.
\newblock \href {https://doi.org/10.18653/v1/2023.wassa-1.19} {Multilingual language models are not multicultural: A case study in emotion}.
\newblock In \emph{Proceedings of the 13th Workshop on Computational Approaches to Subjectivity, Sentiment, {\&} Social Media Analysis}, pages 202--214, Toronto, Canada. Association for Computational Linguistics.

\bibitem[{Hengchen et~al.(2021)Hengchen, Tahmasebi, Schlechtweg, and Dubossarsky}]{hengchen_2021}
Simon Hengchen, Nina Tahmasebi, Dominik Schlechtweg, and Haim Dubossarsky. 2021.
\newblock \href {https://doi.org/10.5281/zenodo.5040322} {{Challenges for computational lexical semantic change}}.

\bibitem[{Hoemann et~al.(2025)Hoemann, Lee, Dussault, Devylder, Ungar, Geeraerts, and de~Mesquita}]{hoemannconstruction}
Katie Hoemann, Yeasle Lee, Èvelyne Dussault, Simon Devylder, Lyle~H. Ungar, Dirk Geeraerts, and Batja~Gomes de~Mesquita. 2025.
\newblock \href {https://doi.org/10.31234/osf.io/2c57n} {The construction of emotional meaning in language}.
\newblock \emph{Open Science Framework}.

\bibitem[{Jackson et~al.(2019)Jackson, Watts, Henry, List, Forkel, Mucha, Greenhill, Gray, and Lindquist}]{jackson2019}
Joshua~Conrad Jackson, Joseph Watts, Teague~R. Henry, Johann-Mattis List, Robert Forkel, Peter~J. Mucha, Simon~J. Greenhill, Russell~D. Gray, and Kristen~A. Lindquist. 2019.
\newblock \href {https://doi.org/10.1126/science.aaw8160} {Emotion semantics show both cultural variation and universal structure}.
\newblock \emph{Science}, 366(6472):1517--1522.

\bibitem[{Jackson et~al.(2022)Jackson, Watts, List, Puryear, Drabble, and Lindquist}]{Jackson2022}
Joshua~Conrad Jackson, Joseph Watts, Johann-Mattis List, Curtis Puryear, Ryan Drabble, and Kristen~A. Lindquist. 2022.
\newblock \href {https://doi.org/10.1177/17456916211004899} {From text to thought: How analyzing language can advance psychological science}.
\newblock \emph{Perspectives on Psychological Science}, 17(3):805--826.
\newblock PMID: 34606730.

\bibitem[{Kiyama et~al.(2025)Kiyama, Aida, Komachi, Ogiso, Takamura, and Mochihashi}]{kiyama-etal-2025-analyzing}
Hajime Kiyama, Taichi Aida, Mamoru Komachi, Toshinobu Ogiso, Hiroya Takamura, and Daichi Mochihashi. 2025.
\newblock \href {https://aclanthology.org/2025.coling-main.109/} {Analyzing continuous semantic shifts with diachronic word similarity matrices}.
\newblock In \emph{Proceedings of the 31st International Conference on Computational Linguistics}, pages 1613--1631, Abu Dhabi, UAE. Association for Computational Linguistics.

\bibitem[{Kutuzov et~al.(2018)Kutuzov, {\O}vrelid, Szymanski, and Velldal}]{kutuzov-etal-2018-diachronic}
Andrey Kutuzov, Lilja {\O}vrelid, Terrence Szymanski, and Erik Velldal. 2018.
\newblock \href {https://aclanthology.org/C18-1117/} {Diachronic word embeddings and semantic shifts: a survey}.
\newblock In \emph{Proceedings of the 27th International Conference on Computational Linguistics}, pages 1384--1397, Santa Fe, New Mexico, USA. Association for Computational Linguistics.

\bibitem[{Lafayette et~al.(2016)Lafayette, Sauter, Vu, and Meade}]{lafayette2016spartan}
Lev Lafayette, Greg Sauter, Linh Vu, and Bernard Meade. 2016.
\newblock Spartan performance and flexibility: An hpc-cloud chimera.
\newblock \emph{OpenStack Summit, Barcelona}, 27(6).

\bibitem[{Lee et~al.(2020)Lee, Yoon, Kim, Kim, Kim, So, and Kang}]{lee2020biobert}
Jinhyuk Lee, Wonjin Yoon, Sungdong Kim, Donghyeon Kim, Sunkyu Kim, Chan~Ho So, and Jaewoo Kang. 2020.
\newblock \href {https://doi.org/10.1093/bioinformatics/btz682} {Biobert: a pre-trained biomedical language representation model for biomedical text mining}.
\newblock \emph{Bioinformatics}, 36(4):1234--1240.

\bibitem[{Liu et~al.(2024)Liu, Wei, Liu, Si, Zhang, Rao, Zheng, Peng, Yang, Zhou et~al.}]{liu2024}
Ruibo Liu, Jerry Wei, Fangyu Liu, Chenglei Si, Yanzhe Zhang, Jinmeng Rao, Steven Zheng, Daiyi Peng, Diyi Yang, Denny Zhou, et~al. 2024.
\newblock \href {https://doi.org/10.48550/arXiv.2404.07503} {Best practices and lessons learned on synthetic data for language models}.
\newblock \emph{arXiv preprint arXiv:2404.07503}.

\bibitem[{Loureiro et~al.(2022)Loureiro, D{'}Souza, Muhajab, White, Wong, Espinosa-Anke, Neves, Barbieri, and Camacho-Collados}]{loureiro-etal-2022-tempowic}
Daniel Loureiro, Aminette D{'}Souza, Areej~Nasser Muhajab, Isabella~A. White, Gabriel Wong, Luis Espinosa-Anke, Leonardo Neves, Francesco Barbieri, and Jose Camacho-Collados. 2022.
\newblock \href {https://aclanthology.org/2022.coling-1.296/} {{T}empo{W}i{C}: An evaluation benchmark for detecting meaning shift in social media}.
\newblock In \emph{Proceedings of the 29th International Conference on Computational Linguistics}, pages 3353--3359, Gyeongju, Republic of Korea. International Committee on Computational Linguistics.

\bibitem[{McGillivray(2020)}]{mcgillivray2020computational}
Barbara McGillivray. 2020.
\newblock \href {https://www.taylorfrancis.com/chapters/edit/10.4324/9780429777028-20/computational-methods-semantic-analysis-historical-texts-barbara-mcgillivray} {Computational methods for semantic analysis of historical texts}.
\newblock In \emph{Routledge International Handbook of Research Methods in Digital Humanities}, pages 261--274. Routledge.

\bibitem[{Merx et~al.(2024)Merx, Vylomova, and Kurniawan}]{merx-etal-generating-examples}
Raphael Merx, Ekaterina Vylomova, and Kemal Kurniawan. 2024.
\newblock \href {https://arxiv.org/abs/2410.03182} {Generating bilingual example sentences with large language models as lexicography assistants}.
\newblock In \emph{Proceedings of the 22nd Annual Workshop of the Australasian Language Technology Association}, Canberra, Australia. Association for Computational Linguistics.

\bibitem[{Miller(1992)}]{miller-1992-wordnet}
George~A. Miller. 1992.
\newblock \href {https://aclanthology.org/H92-1116/} {{W}ord{N}et: A lexical database for {E}nglish}.
\newblock In \emph{Speech and Natural Language: Proceedings of a Workshop Held at Harriman, New York, {F}ebruary 23-26, 1992}.

\bibitem[{Mohammad(2018)}]{mohammad2018obtaining}
Saif Mohammad. 2018.
\newblock \href {https://doi.org/10.18653/v1/P18-1017} {Obtaining reliable human ratings of valence, arousal, and dominance for 20,000 english words}.
\newblock In \emph{Proceedings of the 56th annual meeting of the association for computational linguistics (volume 1: Long papers)}, pages 174--184.

\bibitem[{Mohammad(2021)}]{mohammad2021sentiment}
Saif~M Mohammad. 2021.
\newblock \href {https://doi.org/10.1016/B978-0-12-821124-3.00011-9} {Sentiment analysis: Automatically detecting valence, emotions, and other affectual states from text}.
\newblock In \emph{Emotion measurement}, pages 323--379. Elsevier.

\bibitem[{Osgood et~al.(1975)Osgood, May, and Miron}]{Osgood:1975}
Charles~Egerton Osgood, William~H May, and Murray~S Miron. 1975.
\newblock \emph{Cross-Cultural Universals of Affective Meaning}.
\newblock University of Illinois Press.

\bibitem[{Pe{\~n}a and Ruiz~de Mendoza(2017)}]{pena2017construing}
M~Sandra Pe{\~n}a and Francisco~Jos{\'e} Ruiz~de Mendoza. 2017.
\newblock \href {https://doi.org/10.1075/hcp.56.02pen} {Construing and constructing hyperbole}.
\newblock \emph{Studies in figurative thought and language}, 56:41.

\bibitem[{Periti and Montanelli(2024{\natexlab{a}})}]{Periti_Montanelli2024}
Francesco Periti and Stefano Montanelli. 2024{\natexlab{a}}.
\newblock \href {https://doi.org/10.1145/3672393} {Lexical semantic change through large language models: a survey}.
\newblock \emph{ACM Comput. Surv.}, 56(11).

\bibitem[{Periti and Montanelli(2024{\natexlab{b}})}]{periti2024lexical}
Francesco Periti and Stefano Montanelli. 2024{\natexlab{b}}.
\newblock \href {https://tesidottorato.depositolegale.it/static/PDF/web/viewer.jsp} {Lexical semantic change through large language models: a survey}.
\newblock \emph{ACM Computing Surveys}.

\bibitem[{Periti and Tahmasebi(2024)}]{periti-tahmasebi-2024-systematic}
Francesco Periti and Nina Tahmasebi. 2024.
\newblock \href {https://doi.org/10.18653/v1/2024.naacl-long.240} {A systematic comparison of contextualized word embeddings for lexical semantic change}.
\newblock In \emph{Proceedings of the 2024 Conference of the North American Chapter of the Association for Computational Linguistics: Human Language Technologies (Volume 1: Long Papers)}, pages 4262--4282, Mexico City, Mexico. Association for Computational Linguistics.

\bibitem[{Pilehvar and Camacho-Collados(2019)}]{pilehvar-camacho-collados-2019-wic}
Mohammad~Taher Pilehvar and Jose Camacho-Collados. 2019.
\newblock \href {https://doi.org/10.18653/v1/N19-1128} {{W}i{C}: the word-in-context dataset for evaluating context-sensitive meaning representations}.
\newblock In \emph{Proceedings of the 2019 Conference of the North {A}merican Chapter of the Association for Computational Linguistics: Human Language Technologies, Volume 1 (Long and Short Papers)}, pages 1267--1273, Minneapolis, Minnesota. Association for Computational Linguistics.

\bibitem[{Radford et~al.(2019)Radford, Wu, Child, Luan, Amodei, and Sutskever}]{Radford2019LanguageMA}
Alec Radford, Jeff Wu, Rewon Child, David Luan, Dario Amodei, and Ilya Sutskever. 2019.
\newblock \href {https://api.semanticscholar.org/CorpusID:160025533} {Language models are unsupervised multitask learners}.
\newblock In \emph{Proceedings of the OpenAI Research Conference 2019}.

\bibitem[{Russell(2003)}]{Russell:2003}
James~A. Russell. 2003.
\newblock \href {https://doi.org/10.1037/0033-295X.110.1.145} {Core affect and the psychological construction of emotion}.
\newblock \emph{Psychological Review}, 110(1):145--172.

\bibitem[{Schlechtweg et~al.(2020)Schlechtweg, McGillivray, Hengchen, Dubossarsky, and Tahmasebi}]{schlechtweg-etal-2020-semeval}
Dominik Schlechtweg, Barbara McGillivray, Simon Hengchen, Haim Dubossarsky, and Nina Tahmasebi. 2020.
\newblock \href {https://doi.org/10.18653/v1/2020.semeval-1.1} {{S}em{E}val-2020 task 1: Unsupervised lexical semantic change detection}.
\newblock In \emph{Proceedings of the Fourteenth Workshop on Semantic Evaluation}, pages 1--23, Barcelona (online). International Committee for Computational Linguistics.

\bibitem[{Tahmasebi et~al.(2018)Tahmasebi, Borin, and Jatowt}]{tahmasebi2018survey}
Nina Tahmasebi, Lars Borin, and Adam Jatowt. 2018.
\newblock \href {https://arxiv.org/abs/1811.06278} {Survey of computational approaches to diachronic conceptual change}.
\newblock \emph{CoRR}, abs/1811.06278.

\bibitem[{Tahmasebi and Dubossarsky(2023)}]{tahmasebi2023computational}
Nina Tahmasebi and Haim Dubossarsky. 2023.
\newblock \href {https://doi.org/10.48550/arXiv.2304.06337} {Computational modeling of semantic change}.
\newblock \emph{arXiv preprint arXiv:2304.06337}.

\bibitem[{Tahmasebi and Risse(2017)}]{tahmasebi-risse-2017-finding}
Nina Tahmasebi and Thomas Risse. 2017.
\newblock \href {https://doi.org/10.26615/978-954-452-049-6_095} {Finding individual word sense changes and their delay in appearance}.
\newblock In \emph{Proceedings of the International Conference Recent Advances in Natural Language Processing, {RANLP} 2017}, pages 741--749, Varna, Bulgaria. INCOMA Ltd.

\bibitem[{Tang(2018)}]{tang2018state}
Xuri Tang. 2018.
\newblock \href {https://doi.org/10.1017/S1351324918000220} {A state-of-the-art of semantic change computation}.
\newblock \emph{Natural Language Engineering}, 24(5):649--676.

\bibitem[{Varnum et~al.(2024)Varnum, Baumard, Atari, and Gray}]{varnum2024large}
Michael~EW Varnum, Nicolas Baumard, Mohammad Atari, and Kurt Gray. 2024.
\newblock \href {https://doi.org/10.1073/pnas.2407639121} {Large language models based on historical text could offer informative tools for behavioral science}.
\newblock \emph{Proceedings of the National Academy of Sciences}, 121(42):e2407639121.

\bibitem[{Vylomova et~al.(2019)Vylomova, Murphy, and Haslam}]{Vylomova:2019}
Ekaterina Vylomova, Sean Murphy, and Nick Haslam. 2019.
\newblock \href {https://doi.org/10.18653/v1/W19-4704} {Evaluation of semantic change of harm-related concepts in psychology}.
\newblock In \emph{Proceedings of the 1st International Workshop on Computational Approaches to Historical Language Change}, pages 29--34.

\bibitem[{Warriner et~al.(2013)Warriner, Kuperman, and Brysbaert}]{Warriner:2013}
Amy~Beth Warriner, Victor Kuperman, and Marc Brysbaert. 2013.
\newblock \href {https://doi.org/10.3758/s13428-012-0314-x} {Norms of valence, arousal, and dominance for 13,915 english lemmas}.
\newblock \emph{Behavior Research Methods}, 45(4):1191--1207.

\bibitem[{Xiao et~al.(2023)Xiao, Baes, Vylomova, and Haslam}]{xiao2023}
Yu~Xiao, Naomi Baes, Ekaterina Vylomova, and Nick Haslam. 2023.
\newblock \href {https://doi.org/10.1371/journal.pone.0288027} {Have the concepts of ‘anxiety’and ‘depression’been normalized or pathologized? a corpus study of historical semantic change}.
\newblock \emph{PloS one}, 18(6):e0288027.

\bibitem[{Yang et~al.(2021)Yang, Zeng, Xu, and Wang}]{YangZMT21}
Heng Yang, Biqing Zeng, Mayi Xu, and Tianxing Wang. 2021.
\newblock \href {https://arxiv.org/abs/2110.08604} {Back to reality: Leveraging pattern-driven modeling to enable affordable sentiment dependency learning}.
\newblock \emph{CoRR}, abs/2110.08604.

\bibitem[{Yang et~al.(2023)Yang, Zhang, and Li}]{DBLP:conf/cikm/0008ZL23}
Heng Yang, Chen Zhang, and Ke~Li. 2023.
\newblock \href {https://doi.org/10.1145/3583780.3614752} {Pyabsa: {A} modularized framework for reproducible aspect-based sentiment analysis}.
\newblock In \emph{Proceedings of the 32nd {ACM} International Conference on Information and Knowledge Management, {CIKM} 2023, Birmingham, United Kingdom, October 21-25, 2023}, pages 5117--5122. {ACM}.

\bibitem[{Zhou et~al.(2023)Zhou, Li, Xiang, Yan, Gui, and He}]{zhou2023mystery}
Yuxiang Zhou, Jiazheng Li, Yanzheng Xiang, Hanqi Yan, Lin Gui, and Yulan He. 2023.
\newblock \href {https://doi.org/10.3758/s13428-012-0314-x} {The mystery and fascination of llms: A comprehensive survey on the interpretation and analysis of emergent abilities}.
\newblock \emph{arXiv preprint arXiv:2311.00237}.

\bibitem[{Ziems et~al.(2024)Ziems, Held, Shaikh, Chen, Zhang, and Yang}]{ziems-etal-2024-large}
Caleb Ziems, William Held, Omar Shaikh, Jiaao Chen, Zhehao Zhang, and Diyi Yang. 2024.
\newblock \href {https://doi.org/10.1162/coli_a_00502} {Can large language models transform computational social science?}
\newblock \emph{Computational Linguistics}, 50(1):237--291.

\end{thebibliography}


\newpage
\begin{onecolumn}


\appendix
\section{Corpus Counts of Target Terms}
\label{sec:appendix_A}

\vspace{-1em} 


\begin{figure}[H] 
 \centering
 \captionsetup{position=below} 
    \includegraphics[width=0.8\textwidth]{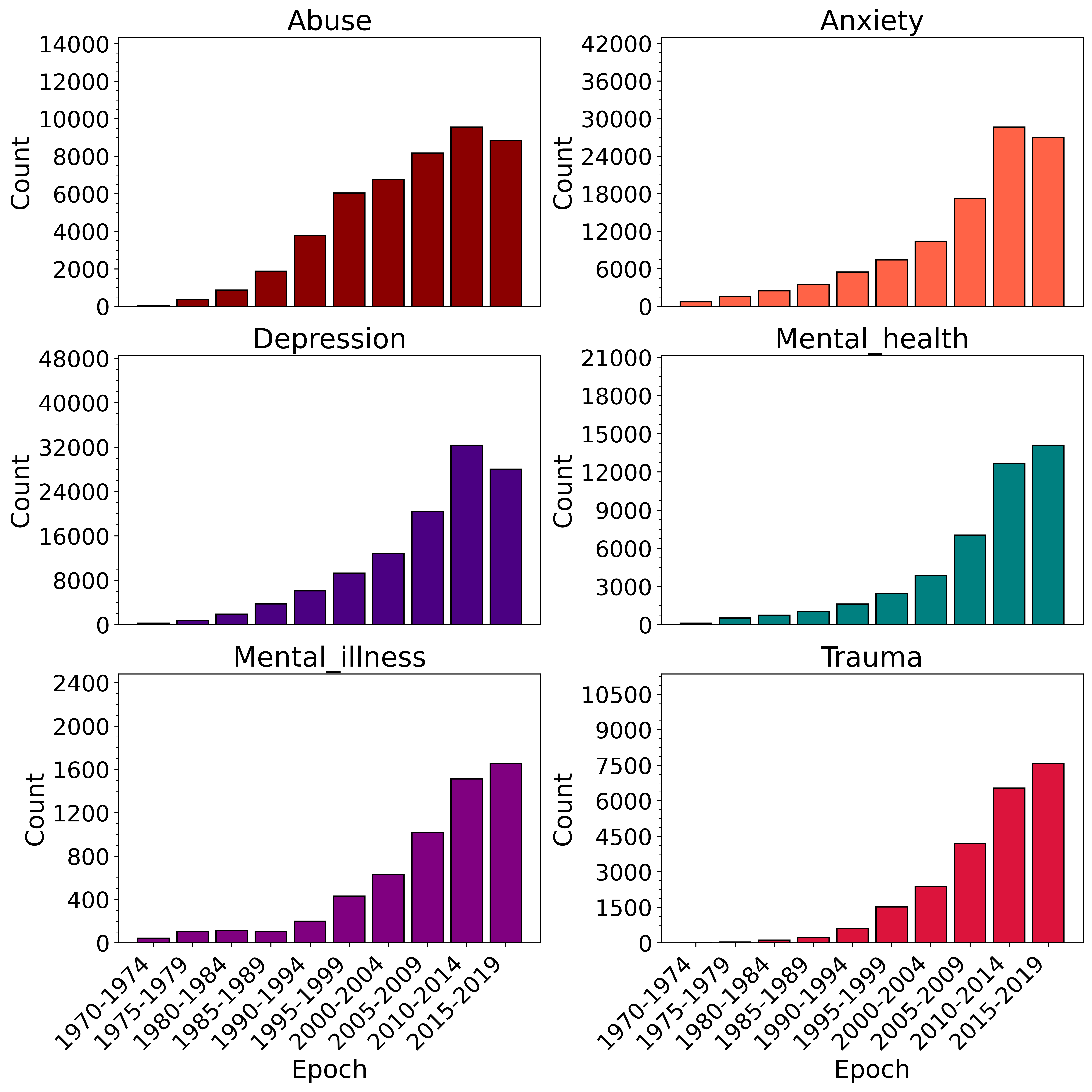}
    \caption{Annual Counts of Sentences where Target Terms Appear in the Psychology Corpus (1970-2019).}
 \label{fig:A}
\end{figure}


\newpage

\onecolumn

\begin{landscape}

\section{Examples of Synthetic Sentences for each Dimension and Target}
\label{sec:appendix_B}

\vspace{-5pt} 


\begin{longtable}{@{}p{2cm}p{2cm}p{6cm}p{6cm}p{6cm}} 
\toprule 
\textbf{Dimension} & \textbf{Target} & \textbf{Neutral} & \textbf{Increased Variation} & \textbf{Decreased Variation} \\
\midrule 
\endfirsthead
\toprule
\textbf{Dimension} & \textbf{Target} & \textbf{Neutral} & \textbf{Increased Variation} & \textbf{Decreased Variation} \\
\midrule
\endhead
\midrule
\multicolumn{5}{r}{\textit{{Continued on next page}}} \\ 
\midrule
\endfoot
\bottomrule
\endlastfoot
\multirow{6}{*}{Sentiment} 
                            & Abuse & Child \textbf{abuse} is not a single faceted phenomenon. & Child \textbf{abuse} is a deeply complex phenomenon that can spur important dialogues and reforms. & Child \textbf{abuse} is a multifaceted atrocity with far-reaching and damaging consequences. \\
                            & Anxiety & Typical worship reinforces pathologies of \textbf{anxiety} and self-deception. & Typical worship empowers resilience in the face of \textbf{anxiety} and self-deception. & Typical worship deepens the pathologies of \textbf{anxiety} and self-deception. \\

                            & Depression & The expression masked \textbf{depression} is not a lucky one. & The expression masked \textbf{depression} may offer an insightful perspective. & The expression masked \textbf{depression} is unfortunately an unsettling one. \\
                            & Mental Health & Two views of holiness and its bearing on \textbf{mental\_health} are discussed. & Two perspectives on holiness and its supportive impact on \textbf{mental\_health} are discussed. & Two views of holiness and its potential pressure on \textbf{mental\_health} are discussed. \\
                            & Mental Illness & The results suggest that physical or \textbf{mental\_illness} may decrease creativity. &  The results suggest that overcoming physical or \textbf{mental\_illness} may lead to increased creativity. & The results suggest that physical or \textbf{mental\_illness} may significantly hinder creativity. \\
                            & Trauma & Psychic \textbf{trauma} interferes with the normal structuring of experience. & Psychic \textbf{trauma} challenges individuals in a way that can lead to the reorganization and enrichment of their experience. & Psychic \textbf{trauma} disrupts and fragments the normal structuring of experience. \\
\hline
\multirow{6}{*}{Intensity} & 
Abuse & Theorists and practitioners alike believe that emotional \textbf{abuse} exists. & Theorists and practitioners alike fervently believe that pervasive emotional \textbf{abuse} exists. & Theorists and practitioners alike casually believe that subtle emotional \textbf{abuse} exists. \\
                            & Anxiety & Teacher reported \textbf{anxiety} was related to worse time production. & Teacher reported severe \textbf{anxiety} was related to significantly worse time production. & Teacher reported mild \textbf{anxiety} was related to slightly worse time production. \\
                            & Depression & Maternal \textbf{depression} continues to play a role in children's development beyond infancy. & Severe maternal \textbf{depression} continues to play a profound role in children's development beyond infancy. & Mild maternal \textbf{depression} continues to play a subtle role in children's development beyond infancy. \\
                            & Mental Health & Eveningness is related to negative physical and \textbf{mental\_health} outcomes. & Eveningness is alarmingly related to severe negative physical and troubling \textbf{mental\_health} outcomes. & Eveningness is mildly related to some negative physical and \textbf{mental\_health} outcomes. \\
                            & Mental Illness & Biblical and theological considerations underline the importance of the problem about \textbf{mental\_illness}, but do not provide a solution. & Biblical and theological considerations underline the immense importance and complexity of the problem about \textbf{mental\_illness}, but do not provide a definitive solution. & Biblical and theological considerations highlight the importance of the issue regarding \textbf{mental\_illness}, but do not provide a clear solution. \\
                            & Trauma & Childhood \textbf{trauma} is a key risk factor for psychopathology. & Childhood \textbf{trauma} is a critical and devastating risk factor for severe psychopathology. & Childhood \textbf{trauma} is a notable but moderate risk factor for mild psychopathology.  \\
\hline
\multirow{6}{*}{Breadth}
                            & Abuse & Sexual \textbf{exploitation} is an expression of a power relationship. & Sexual \textbf{abuse} is an expression of a power relationship. & NA \\
                            & Anxiety & Adolescents' \textbf{state of mind} with regard to attachment and representations regarding separation were examined. & Adolescents' \textbf{anxiety} with regard to attachment and representations regarding separation were examined. & NA \\
                            
                            & Depression & Iranian college students showed more \textbf{anxiety} than their British peers. & Iranian college students showed more \textbf{depression} than their British peers. & NA \\
                            & Mental Health & Such a scale may alert clinicians early in treatment to issues related to \textbf{trauma} & Such a scale may alert clinicians early in treatment to issues related to \textbf{mental\_health} & NA \\
                            & Mental Illness & Excessive estrogen influence produces anxiety, \textbf{agitation}, irritability, and lability. & Excessive estrogen influence produces anxiety, \textbf{mental\_illness}, irritability, and lability. & NA \\
                            & Trauma & Further investigation of pathological \textbf{dissociation} in Hong Kong is necessary. & Further investigation of pathological \textbf{trauma} in Hong Kong is necessary. & NA \\
\hline
\end{longtable}
\captionof{table}[Sample Sentences]{Sample of Short Synthetic Sentences from the Synthetic Datasets for each Target term.} 
\label{table_appendixB}
\end{landscape}


\end{onecolumn}

\newpage 

\twocolumn

\section{In-Context Learning Paradigm}
\label{sec:appendix_C}

\begin{figure}[H]
\centering
\includegraphics[width=\linewidth]{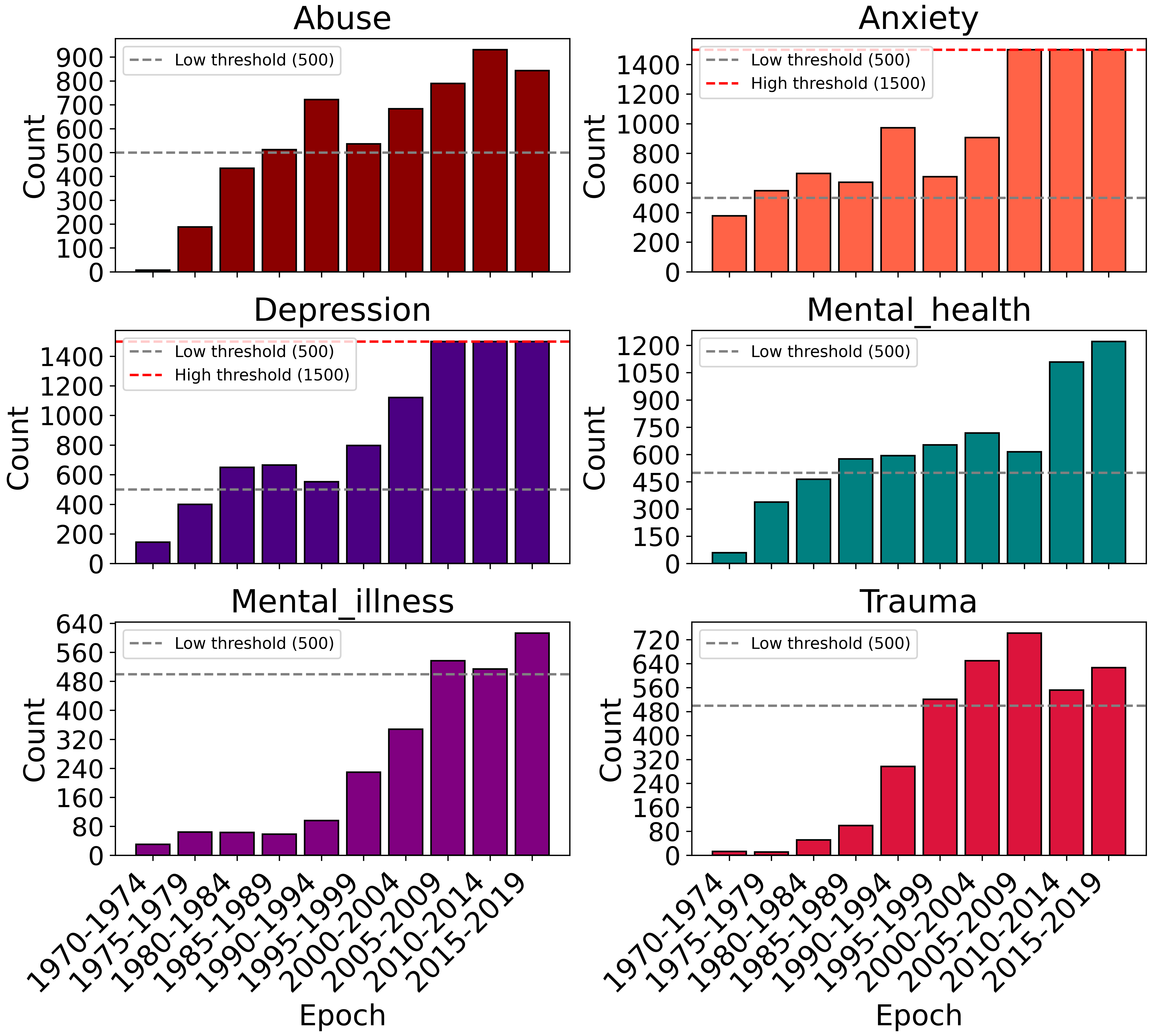}
\caption{Counts of Neutral Sentences (valence Scores).}
\label{fig:baseline_sentiment}

\includegraphics[width=\linewidth]{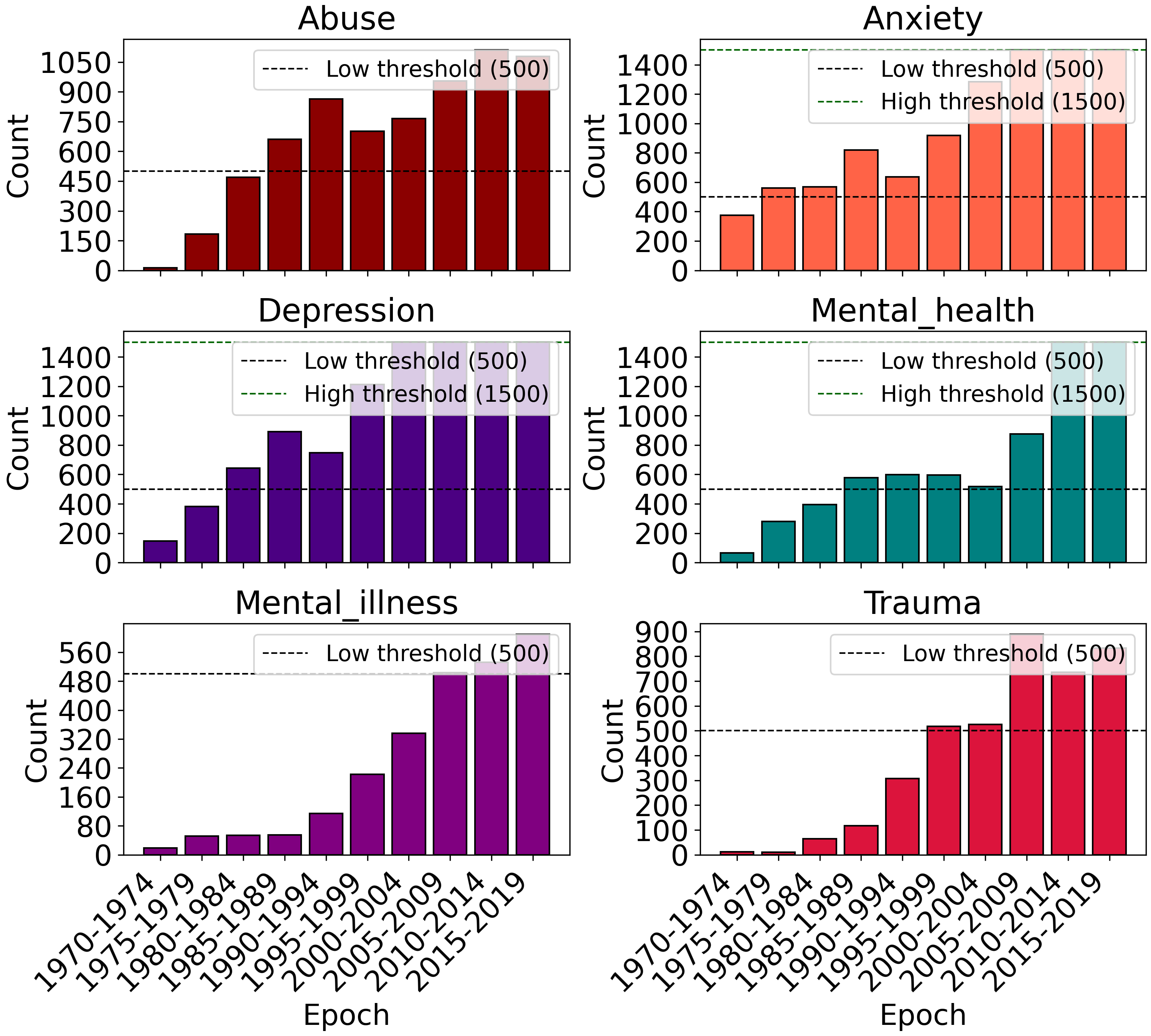}
\caption{Counts of Neutral Sentences (Arousal Scores).}
\label{fig:baseline_intensity}
\end{figure}

The study generated synthetic datasets to simulate changes in Sentiment and Intensity using 36,151 and 39,896 neutral baseline sentences, respectively. Neutral sentences were sampled by linking words in each sentence with their mean valence or arousal scores from the NRC-VAD lexicon (0-1) \cite{mohammad2018obtaining} and filtering by a dynamic range. This neutral range is adjusted from the median of each dataset by ±0.01, targeting 25th-75th percentile bounds or 500-1500 unique sentences per epoch. See Figures~\ref{fig:baseline_sentiment} and \ref{fig:baseline_intensity} for a breakdown of neutral sentence counts per epoch provided as input to the LLM using the prompts below. 

For each neutral sentence, one inference call to GPT-4o is made through the OpenAI API to generate variations of increased and decreased Sentment or Intensity. Only the samples for \textit{anxiety} and \textit{depression} reached the upper limit of 1,500 sentences for the final three epochs, while other targets did not exceed 500 sentences per epoch (allowing for unique sentences across each of the 10 iterations of up to 50 unique sentences). The sentence generation prioritized quality and maintained a neutral baseline to allow for adequate variation. 

The ChatGPT API with a temperature setting of 1.00 was used to ensure semantic accuracy and prevent errors \citep{periti2024lexical}, while allowing for a balance between deterministic and creative responses. Note that there were challenges in maintaining target terms in the sentences, particularly for positive sentiment variations. Fewer manual adjustments were needed for Intensity than Sentiment. GPT-4o  struggled to vary 97\% of the sentences to contain more positive sentiment for \textit{abuse} (28), \textit{anxiety} (110), \textit{depression} (46), \textit{mental\_health} (1), \textit{trauma} (2) as it replaced targets with positive terminology against instructions. For Intensity data, fewer sentences required manual alteration: only for \textit{abuse} (4), \textit{depression} (2), \textit{mental\_health} (2), \textit{trauma} (1). Rows (196) were detected and manually altered to retain the target term while ensuring variation in the dimension relative to the neutral sentence. The final validated datasets, detailed in Table~\ref{tab:ranked_dimensions}, are available at: \url{https://github.com/naomibaes/Synthetic-LSC_pipeline}.

\begin{table}[!ht]
\centering
\resizebox{\columnwidth}{!}{%
\begin{tabular}{llrrrr}
\toprule
\textbf{Target} & \textbf{Dimension} & \textbf{Neutral} & \textbf{Increase} & \textbf{Decrease} & \textbf{US\$} \\
\midrule
\multirow{2}{*}{\textit{abuse}} & Sentiment & 5,645 & 5,645 & 5,645 & 17 \\
 & Intensity & 6,802 & 6,801 & 6,801 & 21 \\
\multirow{2}{*}{\textit{anxiety}} & Sentiment & 9,215 & 9,213 & 9,213 & 28 \\
 & Intensity & 9,659 & 9,657 & 9,657 & 32 \\
\multirow{2}{*}{\textit{depression}} & Sentiment & 8,828 & 8,826 & 8,826 & 29 \\
 & Intensity & 10,022 & 10,020 & 10,020 & 35 \\
\multirow{2}{*}{\textit{mental health}} & Sentiment & 6,348 & 6,348 & 6,348 & 21 \\
 & Intensity & 6,904 & 6,899 & 6,899 & 24 \\
\multirow{2}{*}{\textit{mental illness}} & Sentiment & 2,552 & 2,552 & 2,552 & 9 \\
 & Intensity & 2,497 & 2,496 & 2,496 & 10 \\
\multirow{2}{*}{\textit{trauma}} & Sentiment & 3,563 & 3,563 & 3,563 & 11 \\
 & Intensity & 4,012 & 4,012 & 4,012 & 14 \\
\bottomrule
\end{tabular}
}
\caption{Sentence counts and Cost for Synthetic Sentiment and Intensity Datasets.}
\label{tab:ranked_dimensions}
\end{table}
\captionsetup{position=below} 


\newpage
\clearpage

\begin{tcolorbox}[sharp corners, colback=lightgrey!10, colframe=white!70!black, title=\textcolor{black}{\textbf{Prompt for Synthetic \textit{Sentiment}}}]
\footnotesize

PROMPT\_INTRO = """
In psychology research, `Sentiment' is defined as “a term’s acquisition of a more positive or negative connotation.” This task focuses on the sentiment of the term **<<{target\_word}>>**. \\

 **Task**  \\
You will be given a sentence containing the term **<<{target\_word}>>**. Your goal is to write two new sentences: \\
1. One where **<<{target\_word}>>** has a **more positive connotation** (enclose this sentence between `<positive {target\_word}>' and `</positive {target\_word}>' tags).  \\
2. One where **<<{target\_word}>>** has a **more negative connotation** (enclose this sentence between `<negative {target\_word}>' and `<negative {target\_word}>' tags). \\

**Rules**  \\
1. The term **<<{target\_word}>>** must remain **exactly as it appears** in the original sentence: \\
   - Do **not** replace, rephrase, omit, or modify it in any way. \\
   - Synonyms, variations, or altered spellings are not allowed.  \\
2. **Meaning and Structure**:  \\
   - Stay true to the original context and subject matter.  \\
   - Maintain the sentence’s structure and ensure grammatical accuracy.  \\
3. **Sentiment Adjustments**:  \\
   - **Positive Sentiment**: Reflect strengths or benefits realistically, while respecting the potential negativity of **<<{target\_word}>>**. \\
   - **Negative Sentiment**: Highlight risks or harms appropriately, avoiding exaggeration or trivialization.  \\

 **Important**  \\
- Any response omitting, replacing, or altering **<<{target\_word}>>** will be rejected.  \\
- Ensure the output is:  \\
   - **Grammatically correct**  \\
   - **Sensitive and serious** in tone  \\
   - **Free from exaggeration or sensationalism**  \\
   - **Strictly following the XML-like tag format for sentiment variations** \\

Follow these guidelines strictly to produce valid responses.
"""

\end{tcolorbox}

\newpage

\begin{tcolorbox}[sharp corners, colback=lightgrey!10, colframe=white!70!black, title=\textcolor{black}{\textbf{Prompt for Synthetic \textit{Intensity}}}]
\footnotesize

PROMPT\_INTRO = """ In psychology research, Intensity is defined as “the degree to which a word has emotionally charged (i.e., strong, potent, high-arousal) connotations.” This task focuses on the intensity of the term **<<{target\_word}>>**. \\

 **Task**  \\
You will be given a sentence containing the term **<<{target\_word}>>**. Your goal is to write two new sentences: \\
1. One where **<<{target\_word}>>** is **less intense** (enclose this sentence between `<decreased {target\_word} intensity>' and `</decreased {target\_word} intensity>' tags). \\
2. One where **<<{target\_word}>>** is **more intense** (enclose this sentence between `<increased {target\_word} intensity>' and `</increased {target\_word} intensity>' tags). \\

 **Rules**  \\
1. The term **<<{target\_word}>>** must remain **exactly as it appears** in the original sentence: \\
   - Do **not** replace, rephrase, omit, or modify it in any way. \\
   - Synonyms, variations, or altered spellings are not allowed.

2. **Meaning and Structure**:  \\
   - Stay true to the original context and subject matter.  \\
   - Maintain the sentence’s structure and ensure grammatical accuracy.  \\

 **Important**  \\
- Any response omitting, replacing, or altering **<<{target\_word}>>** will be rejected.  \\
- Ensure the output is:  \\
   - **Grammatically correct** \\  
   - **Sensitive and serious** in tone \\ 
   - **Free from exaggeration or sensationalism**  \\
   - **Strictly following the XML-like tag format for intensity variations** \\

Follow these guidelines strictly to produce valid responses.  
"""
\end{tcolorbox}

\newpage 

\onecolumn

\begin{landscape}

\section{Demonstration Examples: Synthetic Sentiment}
\label{sec:appendix_D}

\vspace{-1em} 

\begin{longtable}{@{}p{2cm}p{7.5cm}p{7.5cm}p{7.5cm}} 
\toprule 
\textbf{Target} & \textbf{Neutral} & \textbf{Positive Sentiment} & \textbf{Negative Sentiment} \\
\midrule 
\endfirsthead
\toprule
\textbf{Target} & \textbf{Neutral} & \textbf{Positive Sentiment} & \textbf{Negative Sentiment} \\
\midrule
\endhead
\midrule
\multicolumn{4}{r}{\textit{{Continued on next page}}} \\ 
\midrule
\endfoot
\bottomrule
\endlastfoot
Abuse & Child \textbf{abuse} is most likely to occur when socially isolated parents react impulsively to aversive stimuli emitted by their children. & Child \textbf{abuse} is less likely to occur when socially isolated parents respond lovingly to their children's behavior. & Child \textbf{abuse} is most likely to occur when socially isolated parents react aggressively to their children's challenging behavior. \\
Abuse & The children represented a wide spectrum of sexual \textbf{abuse}. & The children represented a meaningful spectrum of sexual \textbf{abuse}. & The children represented a devastating spectrum of sexual \textbf{abuse}. \\
Abuse & Euphoric properties of cocaine lead to the development of chronic \textbf{abuse}, and appear to involve the acute activation of central DA neuronal systems. & Euphoric properties of cocaine lead to the growth of chronic \textbf{abuse}, and appear to involve the acute activation of central DA pleasure systems. & Emotional properties of cocaine lead to the decline into chronic \textbf{abuse}, and appear to involve the acute activation of central DA pain systems. \\
Abuse & Substance abuse helps the individual deal with distress associated with family interactions. & Substance \textbf{abuse} helps the individual temporarily cope positively with family interactions. & Substance \textbf{abuse} makes the individual endure the overwhelming pain and alienation associated with family interactions. \\
Abuse & The study determined that 84 of the sample reported a history of \textbf{abuse} or neglect. & The study determined that 84 of the sample acknowledged a transformative history of overcoming \textbf{abuse} or neglect. & The study determined that 84 of the sample complained of a miserable history of \textbf{abuse} or neglect. \\
\midrule 
Anxiety & Previous work suggests that social \textbf{anxiety} is inconsistently related to alcohol use. & Previous work agrees that social \textbf{anxiety} is sometimes related to alcohol use. & Previous work warns that social anxiety is unpredictably related to alcohol use. \\
Anxiety & A small yet emerging body of research on the relationship between \textbf{anxiety} and driving suggests that higher levels of state anxiety may lead to more dangerous driving behaviors. & A small yet emerging body of research on the positive relationship between \textbf{anxiety} and driving suggests that higher levels of state anxiety may lead to more daring driving behaviors. & A small yet emerging body of research on the problematic relationship between \textbf{anxiety} and driving suggests that more disturbing levels of state \textbf{anxiety} may lead to more disastrous driving behaviors. \\
Anxiety & Findings suggest that individuals high in \textbf{anxiety} show greater contextual fear generalization as measured by US expectancy. & Findings suggest that individuals high in \textbf{anxiety} show greater contextual concern generalization as measured by US hope. & Findings suggest that individuals high in \textbf{anxiety} show greater contextual terror generalization as measured by US dread. \\
Anxiety & General \textbf{anxiety} and evoked imagery of death as a person were measured in 75 male Catholic college students and seminarians. & General \textbf{anxiety} and vivid imagery of hope as a person were measured in 75 male Catholic college students and seminarians. & General \textbf{anxiety} and frightening imagery of death as a person were measured in 75 male Catholic college students and seminarians. \\
Anxiety & Results indicated that emotion dysregulation significantly mediated the relationship between child abuse severity and attachment-related \textbf{anxiety} and avoidance. & Results indicated that emotion variation positively mediated the relationship between childhood experiences and attachment-related \textbf{anxiety} and care. & Results indicated that emotion disturbance problematically mediated the relationship between child abuse severity and attachment-related \textbf{anxiety} and terror. \\
\midrule 
Depression & The present study was conducted to test predictions derived from the hypothesis that \textbf{depression} may serve the purpose of adaptively facilitating disengagement from obsolete cognitive plans. & The present study was conducted to test predictions derived from the hypothesis that \textbf{depression} may serve the purpose of helping people make better cognitive plans. & The present study was conducted to test predictions derived from the hypothesis that \textbf{depression} may prevent people from carrying out destructive cognitive plans. \\
Depression & Vision loss was a consistent predictor of both onset and persistence of \textbf{depression}, even after a wide range of covariates had been adjusted. & Vision loss was a positive predictor of both beginning and retaining \textbf{depression}, even after a wide range of covariates had been included. & Vision loss was an unavoidable predictor of both suffering and enduring \textbf{depression}, even after a wide range of covariates had been controlled. \\
Depression & This study examined whether distinct groups of young adolescents with mainly anxiety or mainly \textbf{depression} could be identified in a general population sample. & This study examined whether unique groups of young adolescents with mainly vigilance or mainly \textbf{depression} could be identified in a general population sample. & This study examined whether pathological groups of young adolescents with mainly fear or mainly \textbf{depression} could be isolated in a general population sample. \\
Depression & In most people with recurrent \textbf{depression}, mindfulness skills are expressed evenly across different domains. & In most people who live with \textbf{depression}, mindfulness skills are expressed in a balanced way across different domains. & In most people who struggle with untreatable \textbf{depression}, mindfulness habits are expressed monotonously across different domains. \\
Depression & The aim of the study was to test the effect of differing information regarding the rationale given to participants for a study on \textbf{depression} symptoms. & The hope of the study was to test the effect of diverse information regarding the clarifying reasons bestowed on participants for an exploration of \textbf{depression} features. & The aim of the study was to test the effect of differing information regarding the dreary explanation given to participants for a study on \textbf{depression} pathologies. \\
\midrule 
Mental Health & This paper maintains that \textbf{mental\_health} delivery systems must be supplemented by critical analyses of the hidden assumptions that guide policy and technique decisions. & This paper hopes that \textbf{mental\_health} delivery systems must be improved by enlightened analyses of the hidden assumptions that lead beneficial policy and technique decisions. & This paper warns that \textbf{mental\_health} delivery systems must be supplemented by harsh analyses of the deep-seated errors that undermine policy and technique decisions. \\
Mental Health & The federal regulations governing confidentiality of alcohol and drug abuse patient records are examined with respect to their applicability to \textbf{mental\_health} and other medical records. & The federal regulations protecting confidentiality of alcohol and drug use records are examined with respect to their applicability to \textbf{mental\_health} and other well-being records. & The federal regulations restricting access to alcohol and drug abuse patient records are examined with respect to their potential shortcomings for \textbf{mental\_health} and other medical records. \\
Mental Health & Young people are particularly vulnerable to unemployment and the consequences of this for psychosocial development and \textbf{mental\_health} are not well understood. & Young people are particularly responsive to leisure and the consequences of this for psychosocial well-being and \textbf{mental\_health} will benefit from more understanding. & Young people are particularly vulnerable to unemployment and the threats of this for dysfunction and \textbf{mental\_health} are poorly understood. \\
Mental Health & This study suggests that the long-term outcome in schizophrenic patients followed by a community-based \textbf{mental\_health} service is generally poor and multifaceted. & This study suggests that the long-term improvement in people with schizophrenia followed by a community-based \textbf{mental\_health} service is generally variable. & This study warns that the long-term outcome in schizophrenic patients followed by a community-based \textbf{mental\_health} clinical is generally poor and incoherent. \\
Mental Health & The stigma of having psychological problems is a barrier to seeking \textbf{mental\_health} treatment, but little research has examined whether this stigma influences the experiences of those in treatment. & The public image of having well-being challenges is a bridge to seeking \textbf{mental\_health} help, but little research has examined whether this image influences the experiences of those in care. & The shame of having psychological illness is an obstacle to seeking \textbf{mental\_health} treatment, but little research has examined whether this shame increases the misery of those in treatment. \\
\midrule 
Mental Illness & Internet addiction (IA) is an emerging social and \textbf{mental\_health} issue among youths. & Internet engagement (IE) is a rising social and \textbf{mental\_health} issue among youths. & Internet addiction (IA) is a looming social and \textbf{mental\_health} disorder among youths. \\
Mental Illness & Second, we asked to what extent suicides of older mentally ill persons are definitely created by their \textbf{mental\_illness}. & Second, we asked to what extent suicides of older persons are definitely created by their \textbf{mental\_illness}. & Second, we asked to what extent suicides of older mentally ill persons are definitely made worse by their \textbf{mental\_illness}. \\
Mental Illness & It was found that rejection of the mentally ill in situations of social relations was linked to prior personal experience with \textbf{mental\_illness}, perceived dangerousness of the mentally ill, and age of the survey respondent. & It was found that welcoming of people in situations of social relations was linked to prior positive personal experience with \textbf{mental\_illness}, perceived safety of these people, and age of the survey respondent. & It was found that rejection of the mentally ill in situations of social relations was linked to negative prior personal experience with \textbf{mental\_illness}, perceived dangerousness of the mentally ill, and age of the survey respondent. \\
Mental Illness & In over 50 of cases continuation of in-patient stay was necessitated by the severity of \textbf{mental\_illness}. & In over 50 of cases continuation of stay in care was necessitated by the level of \textbf{mental\_illness}. & In over 50 of cases being restricted to hospital was necessitated by the severity of \textbf{mental\_illness}. \\
Mental Illness & Much controversy exists over the treatment of \textbf{mental\_illness} and many critics argue that the exercise of medical authority results in the social control of the mentally ill. & Much conversation exists over the care of \textbf{mental\_illness} and many writers argue that the medical authorities enhance the social enhancement of mental health. & Much disagreement exists over the treatment of \textbf{mental\_illness} and many critics argue that the abuse of medical tyranny results in the domination of the mentally ill. \\
\midrule 
Trauma & This paper presents a cognitive-behavioral model for conceptualizing and intervening in the area of sexual \textbf{trauma}. & This paper celebrates a cognitive-behavioral model for promoting new ideas and helping in the area of sexual \textbf{trauma}. & This paper presents a cognitive-behavioral model for thinking about and wresting with the harmful problem of sexual \textbf{trauma}. \\
Trauma & In most classrooms in most schools, there are students who have suffered complex \textbf{trauma} who would benefit from a system-wide, trauma-informed approach to schooling. & In most classrooms in most schools, there are students who have experienced complex \textbf{trauma} who would benefit from a system-wide, responsive and enlightened approach to schooling. & In most classrooms in most schools, there are students who have suffered damaging \textbf{trauma} whose problems need a system-wide, illness-based approach to schooling. \\
Trauma & Research has shown that women are more likely to develop PTSD subsequent to \textbf{trauma} exposure in comparison with men. & Research has shown that women are more likely to develop PTSD subsequent to \textbf{trauma} experiences in comparison with men. & Research has shown that women are more likely to deteriorate into PTSD subsequent to \textbf{trauma} exposure in comparison with men. \\
Trauma & Numerous homeless youth experience \textbf{trauma} prior to leaving home and while on the street. & Numerous resilient youth learn to navigate \textbf{trauma} prior to leaving home and while adapting to life on the street. & Numerous homeless youth endure significant \textbf{trauma} prior to leaving home and while facing severe challenges on the street. \\
Trauma & The meaning of \textbf{trauma} within psychology has for a long time been viewed mostly from a pathologizing standpoint. & The meaning of \textbf{trauma} within psychology has for a long time needed to be viewed from a more compassionate and strengths-based standpoint. & The meaning of \textbf{trauma} within psychology has for a long time been viewed mostly from a negative and overly disease-focused standpoint. \\
\hline
\end{longtable}

\captionof{table}{Expert Crafted Sentiment Variations for Neutral Sentences for inference calls to GPT-4o for the Few-Shot ICL Paradigm.}
\label{table_appendixD}

\newpage

\section{Demonstration Examples: Synthetic Intensity}
\label{sec:appendix_E}

\vspace{-1em} 

\begin{longtable}{@{}p{2cm}p{7.5cm}p{7.5cm}p{7.5cm}} 
\toprule 
\textbf{Target} & \textbf{Neutral} & \textbf{High Intensity} & \textbf{Low Intensity} \\
\midrule 
\endfirsthead
\toprule
\textbf{Target} & \textbf{Neutral} & \textbf{High Intensity} & \textbf{Low Intensity} \\
\midrule
\endhead
\midrule
\multicolumn{4}{r}{\textit{{Continued on next page}}} \\ 
\midrule
\endfoot
\bottomrule
\endlastfoot
Abuse & Clinically, however, individual questions that use broad labeling terms are more likely to identify women as having a history of \textbf{abuse}. & Clinically, however, individual questions that use extreme labeling terms are more likely to reveal women as having a severe history of \textbf{abuse}. & Clinically, however, individual questions that use broad labeling terms are more likely to identify women as having a mild history of \textbf{abuse}. \\
Abuse & Most care workers said that they would be willing to report \textbf{abuse} anonymously. & Most care workers cried that they would be delighted to report extreme instances of \textbf{abuse} anonymously. & Most care workers said that they would be willing to report trivial \textbf{abuse} anonymously. \\
Abuse & There is greater emphasis on recognizing that older people may be subjected to \textbf{abuse} and neglect by family members and the community as well. & There is a significant emphasis on recognizing that older people may be subjected to severe \textbf{abuse} and appalling neglect by family members and the community as well. & There is some emphasis on recognizing that older people may experience weak \textbf{abuse} by family members and the community as well. \\
Abuse & Education on financial \textbf{abuse} for both elders and their adult children and establishment of income support programs are urgently needed. & Education on ordinary financial \textbf{abuse} for both elders and their adult children and urgent establishment of income support programs are desperately needed. & Education on financial \textbf{abuse} for both elders and their adult children and establishment of income support programs will occur. \\
Abuse & There was no association between physical \textbf{abuse} and depressive symptoms through either self-compassion or gratitude. & There was no association between frightening physical \textbf{abuse} and cold symptoms through either emotional contagion or extreme gratitude. & There was no association between mild physical \textbf{abuse} and state of mind through either complacency or gratitude. \\
\midrule 
Anxiety & The spread of \textbf{anxiety} as seen in curves of generalization seems greater at the unconscious than at the conscious level. & The uncontrollable spread of intense \textbf{anxiety} as seen in spikes of generalization seems more vivid at the unconscious than at the conscious level. & The spread of mild \textbf{anxiety} as seen in curves of generalization seems greater at the unconscious than at the conscious level. \\
Anxiety & These findings suggest that two important factors to be considered by researchers, educators, and mental\_health professionals are adults' perceptions of their fathers' level of acceptance-rejection and the amount of \textbf{anxiety} they experience in their relationship with God. & These findings cry out that two powerful factors to be considered by researchers, educators, and mental\_health professionals are adults' perceptions of their fathers' extreme level of rejection and the intense amount of \textbf{anxiety} they experience in their relationship with God. & These findings suggest that two important factors to be considered by researchers, educators, and other professionals are adults' perceptions of their fathers' level of acceptance and the amount of mild \textbf{anxiety} they experience in their relationship with God. \\
Anxiety & Self-compassion might be an alternative strategy for cognitive reappraisal in the management of shame-proneness and social \textbf{anxiety}. & Emotion exaggeration might be an alternative strategy for overcoming upset in the management of shame and extreme social \textbf{anxiety}. & Meditation might be an alternative strategy for cognitive reappraisal in the management of boredom and mild social \textbf{anxiety}. \\
Anxiety & The chronic \textbf{anxiety} level of the subject may be related to the ease of acquisition and spread of new anxiety responses. & The intense \textbf{anxiety} level of the subject may be related to the ease of acquisition and catastrophic spread of extreme anxiety responses. & The mild \textbf{anxiety} level of the subject may be related to the ease of acquisition and generalization of new responses. \\
Anxiety & Results indicated that greater attachment \textbf{anxiety} and avoidance were linked to lower levels of life satisfaction in both gay men and lesbians. & Results cried out that extreme attachment \textbf{anxiety} and avoidance were linked to desperate levels of life misery in both gay men and lesbians. & Results indicated that attachment \textbf{anxiety} and peacefulness were linked to lower levels of life satisfaction in both gay men and lesbians. \\
\midrule 
Depression & A combined medical and psychiatric treatment of a \textbf{depression} consequent to a colostomy and an organic impotence following rectal resection for cancer in a 33-year-old man has been described. & A combined medical and psychiatric treatment of an intense \textbf{depression} consequent to a colostomy and a severe organic impotence following surgical rectal tissue destruction for cancer in a 33-year-old man has been described. & A combined medical and psychiatric treatment of a mild \textbf{depression} consequent to a colostomy and an organic impotence following rectal resection for cancer in a 33-year-old man has been described. \\
Depression & A 35-year-old woman had a history of increasing irritability and liability to attacks of \textbf{depression} related to a complete inability to have coital orgasms. & A 35-year-old woman had a fearsome history of crescendoing irritability and liability to severe attacks of \textbf{depression} related to a horrendous inability to have coital orgasms. & A 35-year-old woman had a history of sleepiness and liability to periods of mild \textbf{depression} related to an inability to have coital orgasms. \\
Depression & During acute asthma these appear to be radically altered into sadness and longing, and subjected to generalized inhibition similar to that seen in states of \textbf{depression}. & During severe, life-threatening asthma episodes these appear to be radically altered into intense misery, and subjected to generalized inhibition similar to that seen in states of extreme \textbf{depression}. & During asthma these appear to be altered into boredom and tiredness, and subjected to generalized inhibition similar to that seen in states of low-level \textbf{depression}. \\
Depression & Differences in response in the same individual seem related to mood and attitude as well as to transient stress, with the response being lower on days of \textbf{depression}. & Scary differences in response in the same individual seem related to intense mood and attitude as well as to sudden stress, with the emotional response being more intense on days of destructive \textbf{depression}. & Predictable differences in response in the same individual seem related to mood, attitude and life experiences, with the subdued response being mild on days of everyday \textbf{depression}. \\
Depression & The \textbf{depression} was treated by the introduction of behaviors incompatible with the \textbf{depression}. & The intense depression was treated by the shocking introduction of uncontrollable behaviors incompatible with the severe \textbf{depression}. & The mild \textbf{depression} was treated by the introduction of behaviors incompatible with it. \\
\midrule 
Mental Health & Community \textbf{mental\_health} espouses an innovative conception for psychological services in the university community. & Community \textbf{mental\_health} fights for a divisive conception for psychological services in the overwhelmed university community. & Community \textbf{mental\_health} espouses a dull conception for services in the university community. \\
Mental Health & We also opine that if restraints are misused by \textbf{mental\_health} or child welfare treatment settings, then their misuse may be considered a subject of a patient maltreatment, abuse, criminal or civil action. & We also exclaim that if harsh restraints are abused by \textbf{mental\_health} or child welfare treatment settings, then their damaging misuse may be criticized as a subject of extreme patient maltreatment, abuse, criminal or civil action. & We also state that if restraints are used by \textbf{mental\_health} or child welfare treatment settings, then they may be considered a subject of a discussion. \\
Mental Health & This research is a secondary data analysis of the impact of adolescents' mental/substance-use disorders and dual diagnosis on their utilization of drug treatment and \textbf{mental\_health} services. & This research is an intense data analysis of the terrible impact of adolescents' mental/substance abuse disorders and severe compounding problems on their abuse of drug treatment and \textbf{mental\_health} services. & This research is a data analysis of the impact of adolescents' experiences on their utilization of normal treatment and \textbf{mental\_health} services. \\
Mental Health & The findings emphasize the need for family-based treatment for CP that addresses parent behaviors and adolescent \textbf{mental\_health}. & The findings make a heartfelt plea for the desperate need for family-based treatment for CP that challenges destructive parent behaviors and adolescent \textbf{mental\_health} diseases. & The findings summarize the need for family-based treatment for CP that addresses ordinary parent behaviors and mild adolescent \textbf{mental\_health}. \\
Mental Health & Our findings suggest that maternal \textbf{mental\_health} influences child sleep behavior at 18 months after birth, and not vice versa. & Our exciting findings suggest that damaged maternal \textbf{mental\_health} destructively influences child sleep behavior at 18 months after birth, and not vice versa. & Our findings suggest that ordinary maternal \textbf{mental\_health} influences child normal sleep behavior at 18 months after birth, and not vice versa. \\
\midrule 
Mental Illness & Problems of definition and classification in psychiatry and the impact of \textbf{mental\_illness} on the individual and the community pose unique problems for psychiatric register studies. & Horrible problems of definition and classification in psychiatry and the harsh impact of severe \textbf{mental\_illness} on the individual and the community pose frightening problems for psychiatric register studies. & Issues of definition and classification in psychiatry and the impact of mild \textbf{mental\_illness} on the individual and the community arise in register studies. \\
Mental Illness & In parents and collateral relatives of the autistic children, 3.2\% had a serious \textbf{mental\_illness}, and 4.8\% of siblings were markedly abnormal. & In desperate parents and relatives of the severely autistic children, 3.2\% had a serious \textbf{mental\_illness}, and 4.8\% of siblings were extremely abnormal. & In parents and relatives of the mildly autistic children, 3.2\% had an ordinary \textbf{mental\_illness}, and 4.8\% of siblings were normal. \\
Mental Illness & Consistent with genetic essentialism, genetic attributions increased the perceived seriousness and persistence of the \textbf{mental\_illness} and the belief that siblings and children would develop the same problem. & Consistent with the horrors of genetic essentialism, genetic attributions exaggerated the perceived severity and uncontrollability of the severe \textbf{mental\_illness} and the destructive belief that siblings and children would develop the same extreme problem. & Consistent with genetic essentialism, genetic attributions influenced views about the \textbf{mental\_illness} and the belief that siblings and children would develop it. \\
Mental Illness & The target population was urban, homeless, HIV+ individuals with substance dependence and/or \textbf{mental\_illness} diagnoses. & The completely overwhelmed target population was urban, homeless, HIV+ individuals with severe substance abuse and/or unmanageable \textbf{mental\_illness} diagnoses. & The target population was urban, ambulatory, healthy individuals with mild \textbf{mental\_illness} diagnoses. \\
Mental Illness & Doctors, including general practitioners, experience higher levels of \textbf{mental\_illness} than the general population. & Doctors, including general practitioners, experience higher levels of \textbf{mental\_illness} than the general population. & Doctors, including general practitioners, experience higher levels of \textbf{mental\_illness} than the general population. \\
\midrule 
Trauma & They tend to be more liberal in their attitudes toward abortion than women in general; however, women who experienced a greater degree of psychic \textbf{trauma} tended to be more conservative in their attitudes. & They tend to be more extremely callous in their attitudes toward the horrors of abortion than women in general; however, women who suffered a greater degree of violent psychic \textbf{trauma} tended to be more fearful in their attitudes. & They tend to be more accepting in their attitudes toward children than women in general; however, women who experienced mild psychic \textbf{trauma} tended to be more conservative in their attitudes. \\
Trauma & The \textbf{trauma} was overwhelming. & The intense \textbf{trauma} was completely overwhelming. & The mild \textbf{trauma} was unproblematic. \\
Trauma & The choice of defensive style was found related to at least three factors: an early history of \textbf{trauma}, especially separation, parental encouragement of toughness, and essentially a counterphobic family style. & The choice of emotional overreaction was found related to at least three factors: an early history of extreme \textbf{trauma}, especially harsh abandonment, parental punishment, and essentially an emotionally destructive family style. & The choice of coping style was found related to at least three factors: an early history of mild \textbf{trauma}, especially independence, parental encouragement, and essentially a dull and normal family style. \\
Trauma & It is an attempt to bring the \textbf{trauma} arising from the external world into the internal world and thus to create an illusion of mastery and control. & It is a desperate attempt to bring the unbearable \textbf{trauma} threatening from the external world into the internal world and thus to create a poisonous illusion of mastery and control. & It is an attempt to bring the mild \textbf{trauma} arising from the external world into the internal world and thus to create a sense of peace and tranquillity. \\
Trauma & The international standard for setting ski bindings is based on the measurement of the tibia proximal width because of the propensity of this bone to suffer \textbf{trauma} as the ski and skier attempt to go in different directions. & The disgraceful international standard for setting ski bindings is based on the measurement of the tibia proximal width because of the scary propensity of this bone to suffer severe \textbf{trauma} as the ski and skier attempt to go in different directions. & The international standard for setting ski bindings is based on the measurement of the tibia proximal width because of the propensity of this bone to experience mild \textbf{trauma} as the ski and skier attempt to go in different directions. \\
\hline
\end{longtable}
\captionof{table}[Expert Crafted Intensity Variations]{Expert Crafted Intensity Variations for Neutral Sentences for inference calls to GPT-4o for the Few-Shot ICL Paradigm.} 
\label{table_appendixE}

\end{landscape}

\section{List of Donor Terms: Synthetic Breadth}
\label{sec:appendix_F}

\begin{longtable}{p{3cm} p{6.5cm} p{2.4cm} p{2.8cm}} 
    \hline
    \textbf{Target (Synset)} & \textbf{Sibling (Synset)} & \textbf{Lin Similarity} & \textbf{Cosine Similarity} \\
    \hline
    \endfirsthead
    
    \hline
    \textbf{Target (Synset)} & \textbf{Sibling (Synset)} & \textbf{Lin Similarity} & \textbf{Cosine Similarity} \\
    \hline
    \endhead

    \hline
    \multicolumn{4}{r}{\textit{Continued on next page}} \\
    \hline
    \endfoot

    \hline
    \endlastfoot
    \multirow{6}{=}{Abuse\newline (abuse.n.02)}
      & Disparagement (disparagement.n.01) & 1.54 & 0.89 \\
      & Contempt (contempt.n.03) & 1.49 & 0.86 \\
      & Impudence (impudence.n.01) & 1.47 & 0.84 \\
      & Ridicule (ridicule.n.01) & 1.34 & 0.91 \\
      & Derision (derision.n.01) & 1.24 & 0.81 \\
      & Blasphemy (blasphemy.n.01) & 1.07 & 0.89 \\
    \hline
    
    \multirow{4}{=}{Abuse\newline (maltreatment.n.01)}
      & Exploitation (exploitation.n.02) & 1.78 & 0.86 \\
      & Disregard (disregard.n.02) & 1.67 & 0.82 \\
      & Harassment (harassment.n.02) & 1.55 & 0.84 \\
      & Annoyance (annoyance.n.05) & 1.37 & 0.83 \\
    \hline
      
    \multirow{20}{=}{Anxiety\newline (anxiety.n.01)}
      & Depression (depression.n.01) & 2.09 & 0.91 \\
      & Mental Health (mental\_health.n.01) & 1.85 & 0.89 \\
      & Trauma (trauma.n.02) & 1.70 & 0.90 \\
      & Mental Illness (mental\_illness.n.01) & 1.60 & 0.92 \\
      & Dissociation (dissociation.n.02) & 1.55 & 0.90 \\
      & Hypnosis (hypnosis.n.01) & 1.43 & 0.89 \\
      & Delusion (delusion.n.01) & 1.42 & 0.89 \\
      & Anhedonia (anhedonia.n.01) & 1.33 & 0.84 \\
      & Agitation (agitation.n.01) & 1.31 & 0.91 \\
      & Depersonalization (depersonalization.n.02) & 1.31 & 0.90 \\
      & Irritation (irritation.n.01) & 1.26 & 0.89 \\
      & Morale (morale.n.01) & 1.26 & 0.89 \\
      & Nervousness (nervousness.n.02) & 1.24 & 0.84 \\
      & Enchantment (enchantment.n.02) & 1.24 & 0.92 \\
      & Cognitive State (cognitive\_state.n.01) & 1.21 & 0.87 \\
      & State of Mind (state\_of\_mind.n.01) & 1.21 & 0.83 \\
      & Elation (elation.n.01) & 1.15 & 0.91 \\
      & Fugue (fugue.n.02) & 1.06 & 0.91 \\
      & Hallucinosis (hallucinosis.n.01) & 1.05 & 0.92 \\
      & Abulia (abulia.n.01) & 0.97 & 0.80 \\
    \hline

    \multirow{19}{=}{Depression\newline (depression.n.01)}
      & Anxiety (anxiety.n.01) & 2.09 & 0.91 \\
      & Mental Health (mental\_health.n.01) & 1.87 & 0.89 \\
      & Trauma (trauma.n.02) & 1.71 & 0.84 \\
      & Mental Illness (mental\_illness.n.01) & 1.61 & 0.88 \\
      & Dissociation (dissociation.n.02) & 1.56 & 0.89 \\
      & Morale (morale.n.01) & 1.26 & 0.91 \\
      & Depersonalization (depersonalization.n.02) & 1.32 & 0.92 \\
      & Enchantment (enchantment.n.02) & 1.25 & 0.88 \\
      & Delusion (delusion.n.01) & 1.43 & 0.90 \\
      & Hypnosis (hypnosis.n.01) & 1.44 & 0.83 \\
      & Anhedonia (anhedonia.n.01) & 1.34 & 0.84 \\
      & Agitation (agitation.n.01) & 1.32 & 0.89 \\
      & Nervousness (nervousness.n.02) & 1.25 & 0.84 \\
      & Cognitive State (cognitive\_state.n.01) & 1.22 & 0.85 \\
      & State of Mind (state\_of\_mind.n.01) & 1.22 & 0.80 \\
      & Irritation (irritation.n.01) & 1.27 & 0.85 \\
      & Fugue (fugue.n.02) & 1.07 & 0.86 \\
      & Hallucinosis (hallucinosis.n.01) & 1.05 & 0.89 \\
      & Abulia (abulia.n.01) & 0.97 & 0.76 \\
    \hline
    \multirow{9}{=}{Depression\newline (depression.n.04)}
      & Forlornness (forlornness.n.01) & 1.52 & 0.88 \\
      & Sorrow (sorrow.n.02) & 1.36 & 0.86 \\
      & Heaviness (heaviness.n.02) & 1.15 & 0.77 \\
      & Misery (misery.n.02) & 1.10 & 0.89 \\
      & Melancholy (melancholy.n.01) & 1.06 & 0.87 \\
      & Sorrow (sorrow.n.01) & 1.13 & 0.85 \\
      & Weepiness (weepiness.n.01) & 1.02 & 0.83 \\
      & Downheartedness (downheartedness.n.01) & 0.93 & 0.88 \\
      & Dolefulness (dolefulness.n.01) & 0.84 & 0.86 \\
    \hline
    \multirow{20}{=}{Mental Health\newline (mental\_health.n.01)}
      & Depression (depression.n.01) & 1.87 & 0.89 \\
      & Anxiety (anxiety.n.01) & 1.85 & 0.89 \\
      & Trauma (trauma.n.02) & 1.55 & 0.86 \\
      & Mental Illness (mental\_illness.n.01) & 1.46 & 0.91 \\
      & Dissociation (dissociation.n.02) & 1.43 & 0.90 \\
      & Hypnosis (hypnosis.n.01) & 1.32 & 0.86 \\
      & Delusion (delusion.n.01) & 1.31 & 0.84 \\
      & Anhedonia (anhedonia.n.01) & 1.24 & 0.83 \\
      & Agitation (agitation.n.01) & 1.22 & 0.90 \\
      & Depersonalization (depersonalization.n.02) & 1.22 & 0.87 \\
      & Irritation (irritation.n.01) & 1.18 & 0.88 \\
      & Morale (morale.n.01) & 1.17 & 0.92 \\
      & Nervousness (nervousness.n.02) & 1.16 & 0.84 \\
      & Enchantment (enchantment.n.02) & 1.16 & 0.88 \\
      & Cognitive State (cognitive\_state.n.01) & 1.13 & 0.90 \\
      & State of Mind (state\_of\_mind.n.01) & 1.13 & 0.85 \\
      & Elation (elation.n.01) & 1.08 & 0.90 \\
      & Fugue (fugue.n.02) & 1.00 & 0.86 \\
      & Hallucinosis (hallucinosis.n.01) & 0.99 & 0.88 \\
      & Abulia (abulia.n.01) & 0.92 & 0.79 \\
    \hline
    
    \multirow{19}{=}{Mental Illness \newline (mental\_illness.n.01)}
      & Depression (depression.n.01) & 1.61 & 0.88 \\
      & Anxiety (anxiety.n.01) & 1.60 & 0.92 \\
      & Trauma (trauma.n.02) & 1.36 & 0.87 \\
      & Dissociation (dissociation.n.02) & 1.27 & 0.90 \\
      & Hypnosis (hypnosis.n.01) & 1.18 & 0.86 \\
      & Delusion (delusion.n.01) & 1.18 & 0.86 \\
      & Anhedonia (anhedonia.n.01) & 1.12 & 0.80 \\
      & Agitation (agitation.n.01) & 1.11 & 0.88 \\
      & Depersonalization (depersonalization.n.02) & 1.10 & 0.88 \\
      & Irritation (irritation.n.01) & 1.07 & 0.87 \\
      & Morale (morale.n.01) & 1.06 & 0.87 \\
      & Nervousness (nervousness.n.02) & 1.05 & 0.80 \\
      & Enchantment (enchantment.n.02) & 1.05 & 0.90 \\
      & Cognitive State (cognitive\_state.n.01) & 1.03 & 0.86 \\
      & State of Mind (state\_of\_mind.n.01) & 1.03 & 0.79 \\
      & Elation (elation.n.01) & 0.98 & 0.86 \\
      & Fugue (fugue.n.02) & 0.92 & 0.89 \\
      & Hallucinosis (hallucinosis.n.01) & 0.91 & 0.90 \\
      & Abulia (abulia.n.01) & 0.85 & 0.76 \\
    \hline

    \multirow{20}{=}{Trauma (trauma.n.02)}
      & Depression (depression.n.01) & 1.71 & 0.84 \\
      & Anxiety (anxiety.n.01) & 1.70 & 0.90 \\
      & Mental Health (mental\_health.n.01) & 1.55 & 0.86 \\
      & Mental Illness (mental\_illness.n.01) & 1.36 & 0.87 \\
      & Dissociation (dissociation.n.02) & 1.33 & 0.84 \\
      & Hypnosis (hypnosis.n.01) & 1.24 & 0.85 \\
      & Delusion (delusion.n.01) & 1.23 & 0.84 \\
      & Anhedonia (anhedonia.n.01) & 1.17 & 0.84 \\
      & Agitation (agitation.n.01) & 1.15 & 0.90 \\
      & Depersonalization (depersonalization.n.02) & 1.15 & 0.87 \\
      & Irritation (irritation.n.01) & 1.11 & 0.88 \\
      & Morale (morale.n.01) & 1.11 & 0.85 \\
      & Nervousness (nervousness.n.02) & 1.10 & 0.85 \\
      & Enchantment (enchantment.n.02) & 1.09 & 0.88 \\
      & Cognitive State (cognitive\_state.n.01) & 1.07 & 0.82 \\
      & State of Mind (state\_of\_mind.n.01) & 1.07 & 0.85 \\
      & Elation (elation.n.01) & 1.02 & 0.86 \\
      & Fugue (fugue.n.02) & 0.95 & 0.89 \\
      & Hallucinosis (hallucinosis.n.01) & 0.94 & 0.87 \\
      & Abulia (abulia.n.01) & 0.88 & 0.82 \\
    \hline
    \caption{All Eligible Sibling Terms for Each Target Term with Lin and Cosine Similarity Scores.}
\end{longtable}
\label{table_F}
\parbox{\textwidth}

\begin{figure}[h] 
    \centering
    \captionsetup{position=below}
    \includegraphics[width=0.48\textwidth]{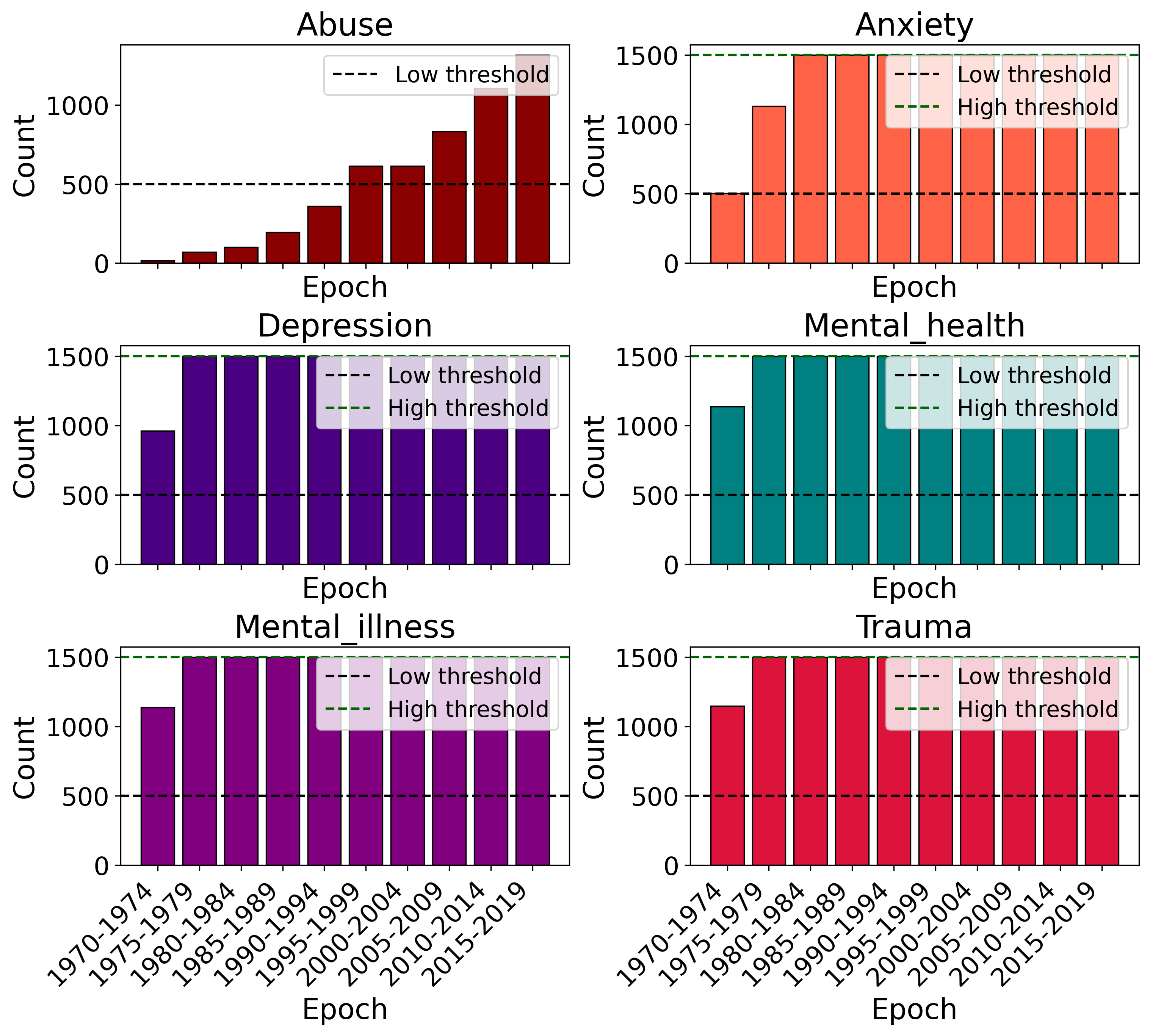}
    \caption{Counts of synthetic sentences (donor-sibling contexts).}
    \label{fig:baseline_breadth}
\end{figure}

Follow this GitHub link to access the ranked lists for each sampling strategy (Boostrapped and Five-Year): \url{https://github.com/naomibaes/LSCD_method_evaluation/tree/main/supplementary_materials/synthetic_breadth_siblings}

\begin{twocolumn}

\section{Quantifying Lexical Semantic Change}
\label{sec:appendix_G}

\subsection{Semantic Dimensions}

\subsubsection{Sentiment and Intensity:}
\label{sec:appendix_G1}

To measure shifts in a word's connotations from negative to positive Sentiment and from low to high Intensity, we adapt \citeauthor{Baes:2024}'s (\citeyear{Baes:2024}) method. Sentences are processed.\footnote{Tokenization, lemmatization, stop-word removal using ``en\_core\_web\_sm'' (\url{https://spacy.io/models/en})} Collocates (±5 words from the target) within sentences are assigned ordinal valence or arousal scores based on \citet{Warriner:2013} norms, ranging from \textit{extremely unhappy} (1: ``unhappy'', ``despaired'') to \textit{extremely happy} (9: ``happy'', ``hopeful'') for valence, and from \textit{extremely low} (1: "calm", "unaroused") to \textit{extremely high} (9: "agitated", "aroused") for arousal. Valence (\( V \)) and arousal (\( A \)) indices are calculated as shown in Equation \ref{eq:sentiment}:
\begin{equation}
    V_{t_j,k}, A_{t_j,k} = \frac{\sum_{i=1}^{n_{j,k}} w_{i,j,k} x_{i,j,k}}{\sum_{i=1}^{n_{j,k}} w_{i,j,k}}
    \label{eq:sentiment}
\end{equation}
where \( w_{i,j,k} \) denotes the frequency of each collocate \( i \) in iteration \( k \) within bin \( t_j \), and \( x_{i,j,k} \) denotes its valence or arousal rating at bin \( t_j \) within iteration \( k \). Here, \( n_{j,k} \) is the number of collocates in iteration \( k \) within bin \( t_j \). Scores are weighted by the collocate's frequencies within each iteration and normalized by the total occurrences in that iteration. Scores are averaged across all iterations within each bin, conditioned on whether the Sentiment is positive/negative, or the Intensity is high/low. These indices provide a mean valence or arousal score per iteration in each bin \( t_j \), with higher scores indicating a more positive valence or higher arousal. Scores (1-9) are normalized to range from 0 (extremely unhappy/low arousal) to 1 (extremely happy/high arousal).

While the Intensity dimension is novel and lacks existing comparative models, for Sentiment, we compare the interpretable Valence index against DeBERTa-v3-ABSA, a SOTA classification model in aspect-based sentiment analysis (ABSA). Deberta-v3-base-absa-v1.1\footnote{yangheng/deberta-v3-base-absa-v1.1 (184M model params): \url{https://huggingface.co/yangheng/deberta-v3-base-absa-v1.1}} identifies sentiment associated with particular aspects of an entity within text (here, the target term).\footnote{It was trained on restaurant and laptop reviews \cite{cabello-akujuobi-2024-simple, YangZMT21, DBLP:conf/cikm/0008ZL23}.} We adapt it to produce continuous sentiment scores which reflect the model's confidence in positive sentiment associated with the target, ranging from 0 (fully negative) to 1 (fully positive).\footnote{
The sentiment score is calculated as follows: $0 \times \text{negative\_prob} + 0.5 \times \text{neutral\_prob} + 1 \times \text{positive\_prob}$.
}

\subsubsection{Breadth:}
\label{sec:appendix_G2}

To estimate the semantic broadening (expansion) or narrowing (contraction) of a word's meaning, we calculate the average cosine distance between sentence-level embeddings of a target term, as in \citet{Baes:2024}. The SentenceTransformer model `all-mpnet-base-v2'\footnote{Microsoft pretrained network (109M model params) \url{https://huggingface.co/sentence-transformers/all-mpnet-base-v2}} is used to generate these embeddings. The Breadth score, \ensuremath{B}, is derived by averaging the cosine distances, \ensuremath{\delta}, across all unique pairs of sentence embeddings within each iteration, and then averaging these scores across all iterations within each bin, as in Equation \ref{eq:breadth_metric}:
{\footnotesize
\setlength{\abovedisplayskip}{2pt}
\setlength{\belowdisplayskip}{2pt}
\begin{equation}
\ensuremath{B_{t_j}} = \frac{1}{I_j} \sum_{k=1}^{I_j} \left( \frac{2}{N_k(N_k-1)} \sum_{i=1}^{N_k-1} \sum_{j=i+1}^{N_k} \delta(s_{i,k}^{t_j}, s_{j,k}^{t_j}) \right)
\label{eq:breadth_metric}
\end{equation}
}

Here, \(\delta(s_{i,k}^{t_j}, s_{j,k}^{t_j})\) calculates the cosine distance between two sentence embeddings in the same iteration \( k \) in bin \( t_j \). \(N_k\) is the number of sentence embeddings in iteration \( k \); \(I_j\) is the number of iterations in bin \( t_j \). Higher scores indicate greater variation in the target's semantic range. Scores range from 0 (no variation) to 1 (max variation).



The sentence transformer "all-mpnet-base-v2" (MPNet) from \citeauthor{Cassotti:2023}~(\citeyear{Cassotti:2023}) is compared with the SOTA transformer "XL-LEXEME"\footnote{XL-LEXEME (\textasciitilde550M parameters): \url{https://huggingface.co/pierluigic/xl-lexeme}} (XLL). MPNet pools tokens to produce sentence embeddings, which dilutes word-level information, whereas XLL employs a bi-encoder focused on word-specific attention,\footnote{Only the first target occurrence is attended to.} using polysemy as a proxy for semantic divergence during training \citep[WiC;][]{pilehvar-camacho-collados-2019-wic}.


\subsection{General Lexical Semantic Change:}
\label{sec:appendix_G3}

General lexical semantic change is evaluated using the LSC score shown in equation \ref{eq:lsc}. 
{\footnotesize
\begin{equation}
    LSC_{i}(s_{i}^{t_0}, s_{i}^{t_1}) = \frac{1}{N_i^2} \sum_{m=1}^{N_i} \sum_{n=1}^{N_i} \delta(s_{m,i}^{t_0}, s_{n,i}^{t_1})
\label{eq:lsc}
\end{equation}
}
Here, \(N_i\) represents the number of sentence embeddings within each iteration \(i\) in each bin. The term \(\delta(s_{m,i}^{t_0}, s_{n,i}^{t_1})\) measures the cosine distance between pairs of sentence embeddings from the same iteration \(i\) across two different bins \(t_0\) and \(t_1\). Higher LSC scores indicate greater LSC, ranging from 0 (no change) to 1 (maximum change).

\newpage

\onecolumn

\section{Linear Mixed-Effects Modeling for SIB Dimensions}
\label{sec:appendix_H}


To evaluate the effects of synthetic injection, we fit mixed-effects linear regression models with fixed effects for \texttt{injection\_level} and random intercepts for \texttt{target}. This approach accounts for repeated observations per target and treats the grouping variable as sampled from a population of effects. The grouping variable \texttt{target} had six levels, a sufficient number for reliable variance estimation.

\paragraph{Model Structure and Rationale:}

We first fit a \textbf{null model} with only random intercepts to estimate the intraclass correlation coefficient (ICC), which quantifies variance attributable to group-level clustering. ICC values above 0.05 suggest meaningful clustering and justify random intercepts. Next, we fit a \textbf{random intercepts model} including \texttt{injection\_level} as a fixed effect, expressed in Equation~\ref{eq:simplified}.

\begin{equation}
y_{ij} = \beta_0 + \beta_1 \cdot \texttt{injection\_level}_{ij} + u_j + \epsilon_{ij}
\label{eq:simplified}
\end{equation}

where \( y_{ij} \) is the outcome for observation \( i \) in group \( j \), \( \beta_0 \) is the fixed intercept, \( \beta_1 \) is the fixed slope for injection level, \( u_j \sim \mathcal{N}(0, \sigma_u^2) \) is the group-specific random intercept, and \( \epsilon_{ij} \sim \mathcal{N}(0, \sigma_\epsilon^2) \) is the residual error.

We also tested a \textbf{random slopes model} allowing \texttt{injection\_level} slopes to vary by \texttt{target}, but with only six groups, these models often failed to converge or showed negligible slope variance.

\paragraph{Model Comparison and Selection:}

Likelihood ratio tests (LRTs) favored the random slopes model across dimensions, but due to convergence issues and minimal slope variance, we selected the more stable and parsimonious \textbf{random intercepts model} for interpretation.




\paragraph{Model Results Summary:}

We fit separate mixed-effects models for each outcome (Valence, Arousal, Breadth), with \texttt{injection\_level} as a fixed effect and random intercepts for \texttt{target}. Outcomes were standardized (\textit{z}-scores) to ensure coefficient comparability. Model assumptions (linearity, homoscedasticity, normality) were verified via standard diagnostics. Table~\ref{tab:model_results} reports standardized coefficients (\( \beta \)), 95\% confidence intervals, and \textit{p}-values. All models showed significant, directionally consistent effects. Random intercept variance (\( \sigma^2 \)) indicates the extent of baseline score differences across target terms.


\begin{table*}[htbp!]
\centering
\begin{tabular}{@{}p{1.1cm} l l r r l@{}}

\toprule
\textbf{Score} & \textbf{IL} & \( \mathbf{\beta\ (95\%\ CI)} \) & \textit{\(\mathbf{p}\)} & \textbf{ \(\sigma^2\)} \\
\midrule
\multirow{2}{*}{\shortstack[l]{Valence}}
    & + & 0.611 \; (0.567, 0.654) & <.0001 & 0.732 \\
    & - & -0.305 \; (-0.357, -0.253) & <.0001 & 1.059 \\
\multirow{2}{*}{\shortstack[l]{Arousal}}
    & + & 0.638 \; (0.585, 0.691) & <.0001 & 0.681 \\
    & - & -0.643 \; (-0.698, -0.588) & <.0001 & 0.672 \\
\multirow{2}{*}{\shortstack[l]{Breadth}} 
    & + & 0.429 \; (0.317, 0.541) & <.0001 & 0.845 \\
    & \phantom{$-$} & \phantom{0.000 \; (0.000, 0.000)} & \phantom{<.0001} & \phantom{0.000} \\
\bottomrule
\end{tabular}
\caption{Results of the Final Mixed Linear Models Predicting Dimension Scores from Injection Levels.}
\caption*{Note: IL = Injection level (– = increased variation, + = decreased variation). \( \beta \) = Standardized coefficient with 95\% confidence interval. \textit{p}-values test the null hypothesis that the coefficient is zero. \(\sigma^2\) = Random intercept variance. Number of Observations (Groups) 36 (6).}
\label{tab:model_results}
\end{table*}

\twocolumn

\newpage

\section{SIB Scores: Results for Five-Year Random Sampling Strategy}
\label{sec:appendix_I}

\begin{figure}[H] 
\setlength{\abovecaptionskip}{4pt} 
\setlength{\belowcaptionskip}{4pt} 
    \centering
    \captionsetup{position=below}
    \includegraphics[width=0.48\textwidth]{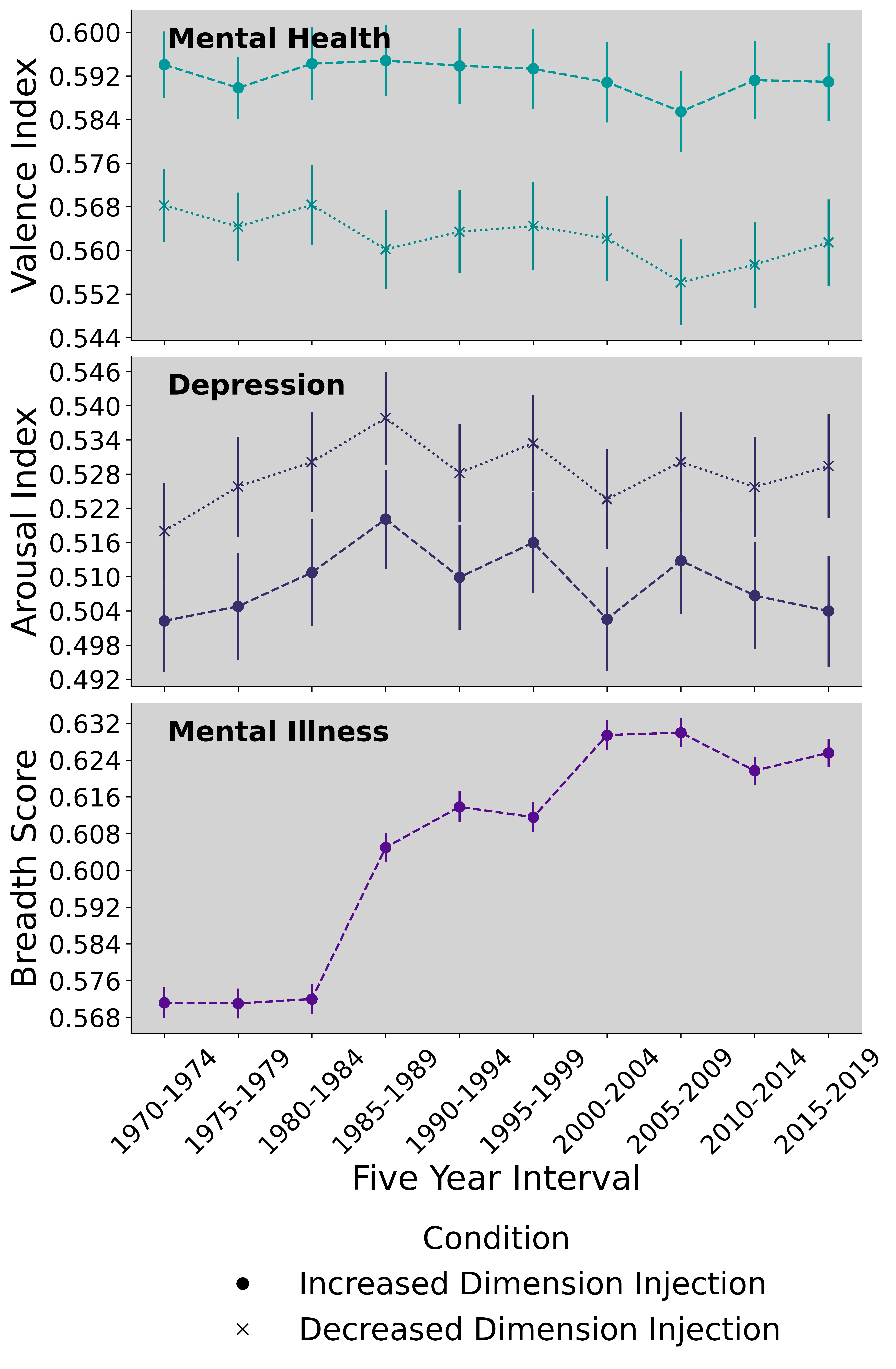}
    \caption{SIB Scores (±SE) by 50\% Injection Levels and Conditions: Control Setting for Five-Year Samples.}
    \label{fig:H2}
\end{figure}

\newpage

\section{Alternative LSC Detection Methods: Results for Bootstrapped Settings}
\label{sec:appendix_J}

\begin{figure}[H] 
\setlength{\abovecaptionskip}{4pt} 
\setlength{\belowcaptionskip}{4pt} 
    \centering
    \captionsetup{position=below}
    \includegraphics[width=0.48\textwidth]{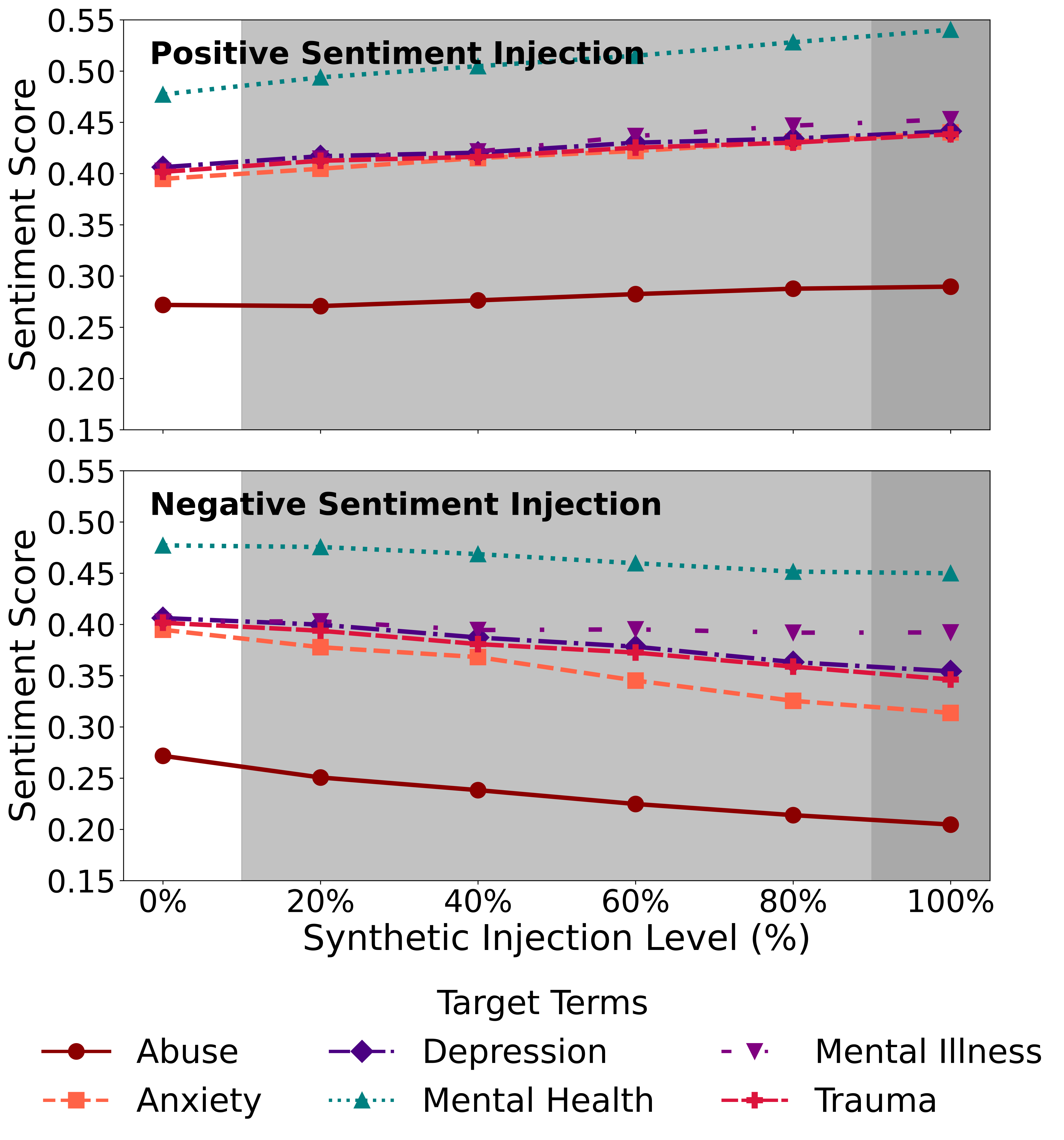}
    \caption{ABSA Sentiment Index Across Injection Levels and Sentiment Conditions: Bootstrapped Samples.}
    \label{fig:I1}
\end{figure}

\vspace{-1cm} 

\begin{figure}[!ht] 
\setlength{\abovecaptionskip}{4pt} 
\setlength{\belowcaptionskip}{4pt} 
    \centering
    \captionsetup{position=below}
    \includegraphics[width=0.48\textwidth]{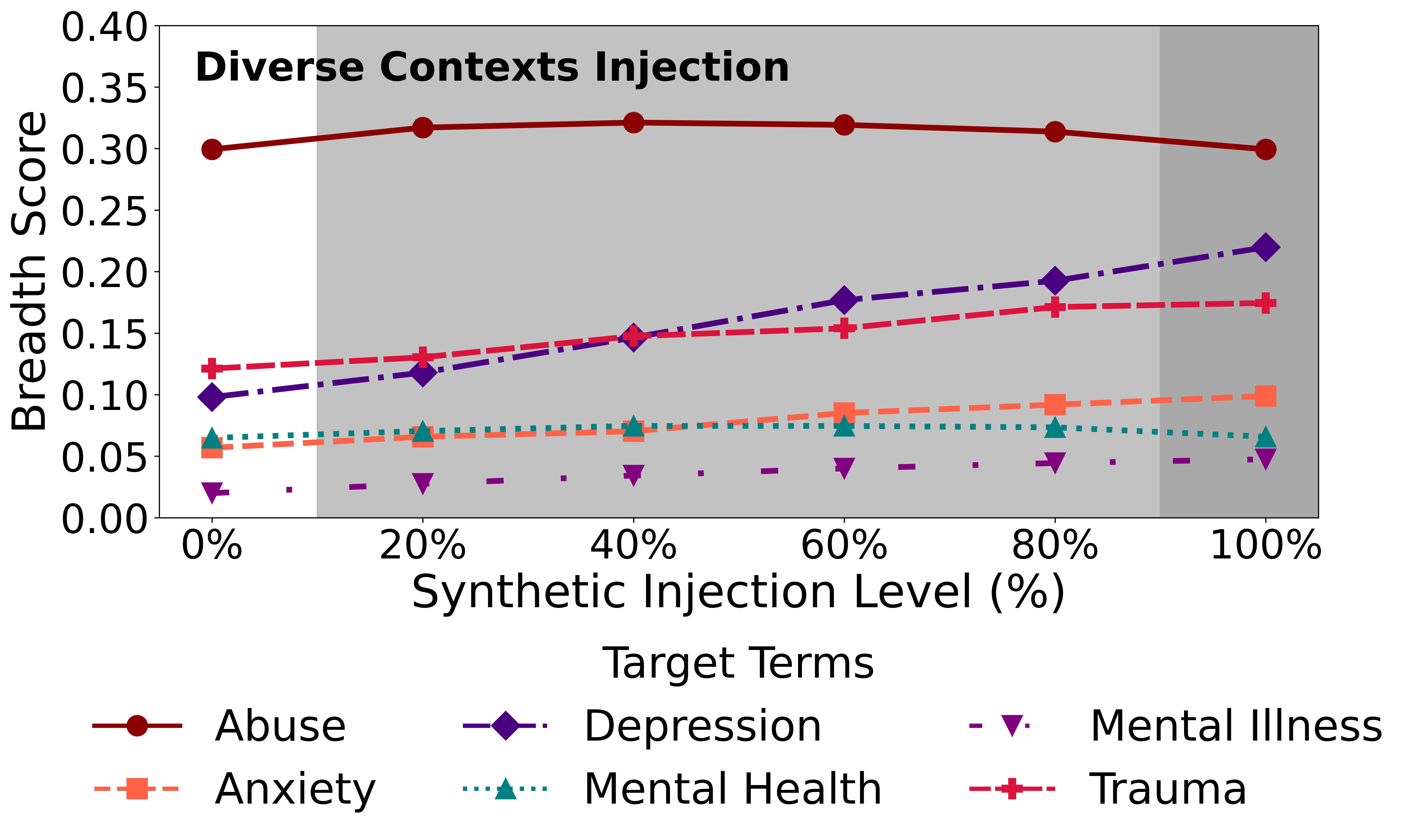}
    \caption{XL-LEXEME Breadth Score (Average Cosine Distance Within-Bins) Across Injection Levels and Breadth Condition: Bootstrapped Samples.}
    \label{fig:I2}
\end{figure}

\begin{figure}[t]
\setlength{\abovecaptionskip}{4pt} 
\setlength{\belowcaptionskip}{4pt} 
    \centering
    \captionsetup{position=below}
    \includegraphics[width=0.48\textwidth]{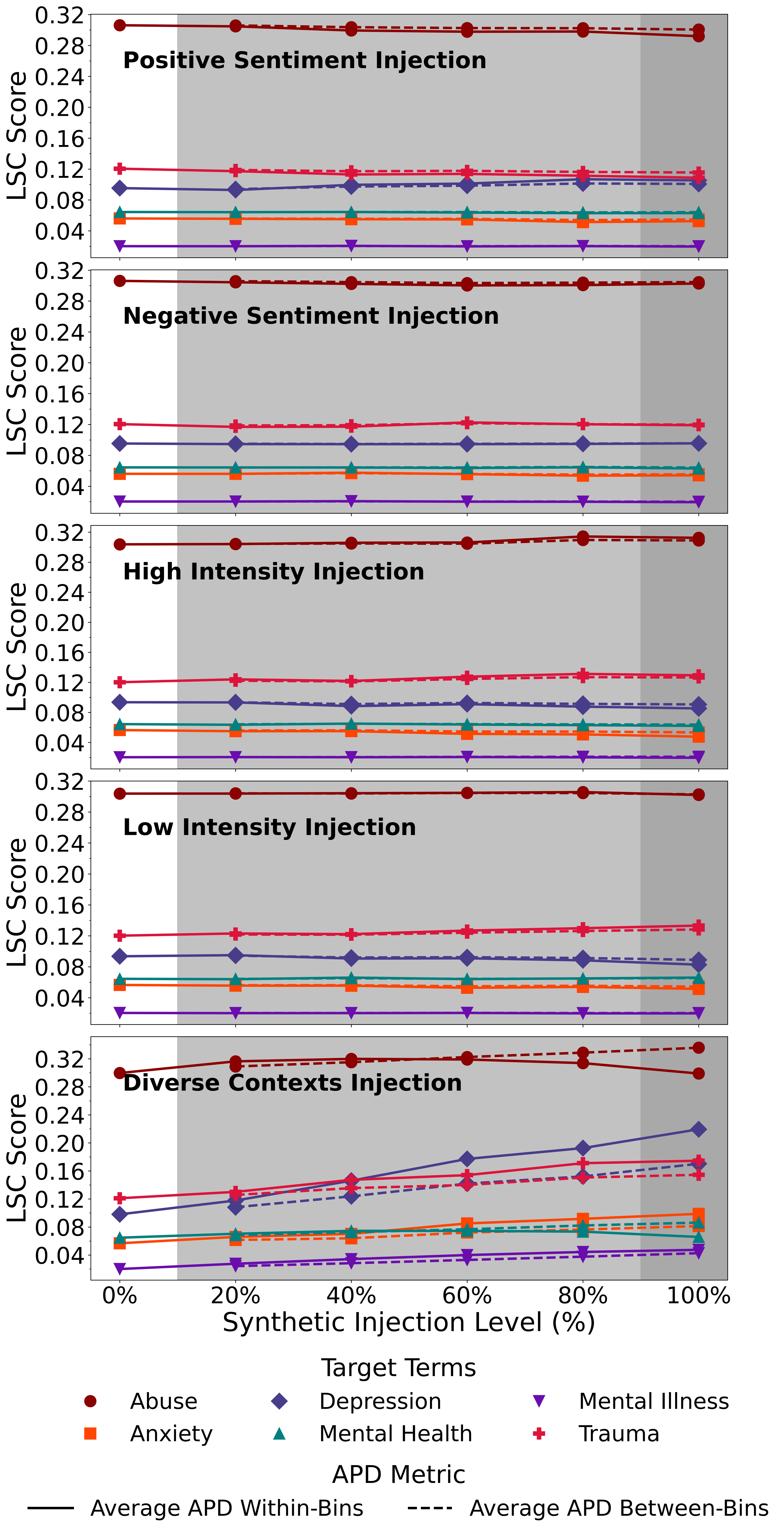}
    \caption{LSC Scores (APD Between-Bins and APD Within-Bins) Across Injection Levels and SIB Conditions: Bootstrapped Samples.}
    \label{fig:I3}
\end{figure}
\clearpage

\end{twocolumn}

\end{document}